\documentclass[lettersize,journal]{IEEEtran}
\IEEEoverridecommandlockouts

\usepackage{bbm}
\usepackage{url}
\usepackage{tikz}
\usepackage{xurl}
\usepackage{array}
\usepackage{xfrac}
\usepackage{bbding}
\usepackage{amsthm}
\usepackage{amsmath}
\usepackage{multirow}
\usepackage{makecell}
\usepackage{pgfplots}
\usepackage{booktabs}
\usepackage{multicol}
\usepackage{hyperref}
\usepackage{threeparttable}
\usepackage{color,xcolor,colortbl}
\usepackage{amsmath,amssymb,amsfonts}
\usepackage{graphicx}
\usepackage{subfigure}
\usepackage{cases}
\usepackage{lscape,lipsum}

\usepackage{cuted}
\usepackage{amsthm}
\usepackage{extarrows}

\usepackage{tikz}
\usetikzlibrary{tikzmark,calc,matrix,positioning,arrows.meta}
\usetikzlibrary{decorations.pathreplacing}

\newtheorem{lemma}{Lemma}

\newtheorem{theorem}{Theorem}

\newtheorem{corollary}{Corollary}
\newtheorem{definition}{Definition}

\newtheorem{observation}{Observation}
\newtheorem{proposition}{Proposition}
\newtheorem{keyquestion}{Question}

\usepackage{cite}
\usepackage{hyperref}
\hypersetup{
    colorlinks,
    linkcolor={red!70!black},
    citecolor={blue!70!black},
    urlcolor={blue!70!black}
}

\usepackage{amsmath,amssymb,amsfonts}
\usepackage{algorithmic}
\usepackage{graphicx}
\usepackage{textcomp}
\usepackage{xcolor}
\def\BibTeX{{\rm B\kern-.05em{\sc i\kern-.025em b}\kern-.08em
    T\kern-.1667em\lower.7ex\hbox{E}\kern-.125emX}}
\begin{document}

\title{Governing AI Forgetting:\\Auditing for Machine Unlearning Compliance}

\author{Qinqi Lin, Ningning Ding, Lingjie Duan,~\IEEEmembership{Senior Member,~IEEE,} Jianwei Huang,~\IEEEmembership{Fellow,~IEEE}
\IEEEcompsocitemizethanks{
Qinqi Lin is with the School of Science and Engineering, Shenzhen Institute of Artificial Intelligence and Robotics for Society, The Chinese University of Hong Kong, Shenzhen, China (e-mail: qinqilin@link.cuhk.edu.cn). Ningning Ding is with the Data Science and Analytics Thrust, Information Hub, Hong Kong University of Science and Technology (Guangzhou), China (e-mail: ningningding@hkust-gz.edu.cn). Lingjie Duan is with the Internet of Things Thrust, Information Hub, Hong Kong University of Science and Technology (Guangzhou), China (e-mail: lingjieduan@hkust-gz.edu.cn). Jianwei Huang is with the School of Science and Engineering, Shenzhen Institute of Artificial Intelligence and Robotics for Society, The Chinese University of Hong Kong, Shenzhen, China (\textit{corresponding author}, e-mail: jianweihuang@cuhk.edu.cn).
}\vspace{-15pt}
}

\maketitle

\markboth{IEEE Transactions on Mobile Computing}%
{Shell \MakeLowercase{\textit{et al.}}: Bare Demo of IEEEtran.cls for Computer Society Journals}

\begin{abstract}
Despite legal mandates for the right to be forgotten, AI operators routinely fail to comply with data deletion requests. While machine unlearning (MU) provides a technical solution to remove personal data's influence from trained models, ensuring compliance remains challenging due to the fundamental gap between MU's technical feasibility and regulatory implementation. In this paper, we introduce the first economic framework for auditing MU compliance, by integrating certified unlearning theory with regulatory enforcement. We first characterize MU's inherent verification uncertainty using a hypothesis-testing interpretation of certified unlearning to derive the auditor's detection capability, and then propose a game-theoretic model to capture the strategic interactions between the auditor and the operator. A key technical challenge arises from MU-specific nonlinearities inherent in the model utility and the detection probability, which create complex strategic couplings that traditional auditing frameworks do not address and that also preclude closed-form solutions. We address this by transforming the complex bivariate nonlinear fixed-point problem into a tractable univariate auxiliary problem, enabling us to decouple the system and establish the equilibrium existence, uniqueness, and structural properties without relying on explicit solutions. Counterintuitively, our analysis reveals that the auditor can optimally reduce the inspection intensity as deletion requests increase, since the operator's weakened unlearning makes non-compliance easier to detect. This is consistent with recent auditing reductions in China despite growing deletion requests. Moreover, we prove that although undisclosed auditing offers informational advantages for the auditor, it paradoxically reduces the regulatory cost-effectiveness relative to disclosed auditing. Experimental results based on real data show that, compared to the state-of-the-art benchmark, disclosed auditing increases the auditor's payoff by up to $2549.30\%$ and the operator's payoff by up to $74.60\%$.
\end{abstract}


\section{Introduction}

\subsection{Background and Motivations}
\IEEEPARstart{A}{s} artificial intelligence (AI) systems become increasingly integrated into everyday applications, concerns over personal data privacy have grown, with 57\% of global consumers viewing AI's handling of personal data as a major threat~\cite{fazlioglu_2023_iapp}. In response, regulations such as the EU's General Data Protection Regulation (GDPR) \cite{gdpr_2018_general} and China's~Personal~Information~Protection Law (PIPL) \cite{thena} have earlier introduced the ``right to be forgotten'' (RTBF), obligating AI operators to delete personal data upon request.\footnote{We adopt the term ``AI operator'' from the EU AI Act~\cite{euartificialintelligenceact_2021}, which broadly refers to any provider, deployer, or authorized representative of AI systems.} Since its introduction, RTBF has triggered millions of deletion requests, including 3.2 million to Google between 2014 and 2019 \cite{bertram2019five}. However, ensuring data deletion compliance in the AI era remains challenging: simply deleting data from storage is insufficient, as trained models still retain their influence, undermining user privacy~\cite{nguyen2022survey}.

To address the challenge of data retention in trained models, \textit{machine unlearning} (MU) has emerged as a promising technical solution~\cite{cao2015towards}. MU attempts to remove or bound the influence of specific data from a model without the prohibitive cost of retraining from scratch. Since its inception, the academic community has developed increasingly efficient MU methods (e.g., \cite{bourtoule2021machine, guo2019certified, sekhari2021remember, qiao2024hessian}), and companies such as Hirundo \cite{a2025_hirundo} are advancing industrial MU solutions for real-world applications. Building on these advances, policymakers, including the European Data Protection Board, are now investigating the integration of MU into emerging data and AI governance frameworks to support data-deletion compliance~\cite{shrishak_2024_support}.

Despite its technical promise, MU faces a fundamental \textit{enforcement challenge}: AI operators may be reluctant to implement it voluntarily for data-deletion compliance~\cite{weng2024proof}. In reality, non-compliance with data-deletion requests is widespread; for example, Everalbum~\cite{a2021_california} and Clearview AI~\cite{lomas_2024_clearview} retained personal data even after explicit deletion requests. This compliance gap is particularly pronounced for MU due to two factors. First, the opacity of AI models and MU procedures creates information asymmetry: without privileged access, users cannot ascertain whether their data has been unlearned, allowing operators to ignore deletion requests without accountability~\cite{zhang2024verification}. Second, MU presents economic disincentives: it reduces model accuracy and generalization~\cite{sekhari2021remember}, lowering system profitability and the commercial value of AI models~\cite{wang2025unlearning}. Consequently, AI operators have little incentive to implement MU voluntarily.

To bridge this compliance gap, regulators are increasingly adopting \textit{AI auditing} as the core enforcement mechanism \cite{a2024_ai,a2020_guidance}. For example, under the GDPR and the EU AI Act~\cite{euartificialintelligenceact_2021}, EU Data Protection Authorities (DPAs) are empowered to inspect AI systems and impose penalties for non-compliance~\cite{_2024_statement}.~In parallel, researchers are developing MU verification techniques (e.g.,~\cite{zhang2024verification}) to equip auditors with technical tools. Together, authorized inspections mitigate information asymmetry through independent verifications, whereas penalty schemes counteract economic disincentives to deter non-compliance. Despl regulatory desite these emerging legislative and technical developments, the principles guiding optimaign for auditing MU compliance remain largely unexplored. This research thus seeks to answer the following two fundamental questions:

\begin{keyquestion}
How will the AI operator best manage MU's data deletion effectiveness when facing non-compliance risks under regulatory audits?
\end{keyquestion}
\begin{keyquestion}
How should the auditor respond to the AI operator's strategic non-compliance and design inspection policies to enforce MU compliance cost-effectively?
\end{keyquestion}

\begin{figure}[t]
    \centering
    \includegraphics[width=\linewidth]{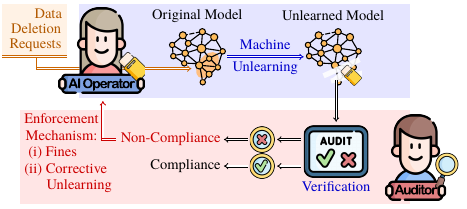}
    \vspace{-15pt}
    \caption{Framework for Auditing Machine Unlearning (MU) Compliance.} 
    \vspace{-15pt}
    \label{fig:illustration_framework}
\end{figure} 

Fig.~\ref{fig:illustration_framework} illustrates this interaction: upon receiving data deletion requests, the AI operator implements MU while the auditor conducts compliance audits to verify deletion effectiveness and impose penalties for detected non-compliance. Answering these questions can help understand this interaction to bridge the fundamental gap between MU's technical capabilities and regulatory implementations, and provide practical insights into regional differences in AI auditing practices.
\vspace{-10pt}
\subsection{Challenges}
In addressing these questions, we face three key challenges arising from the unique nature of MU compliance auditing.

\textit{(1) The first major challenge lies in the inherent uncertainty of MU verification.} 
Conventional data deletion (e.g., database erasure \cite{sarkar2020lethe}) allows deterministic verification through database queries or search-index checks. In contrast, MU techniques are inherently statistical~\cite{sekhari2021remember}, permitting only probabilistic verification of data deletion. This inherent uncertainty complicates regulatory enforcement in two ways: existing MU-verification implementations (e.g.,~\cite{sommer2020towards}) lack a theoretical quantification of the auditor's detection capability, and probabilistic guarantees fundamentally alter the auditor-operator strategic interactions compared to deterministic settings.

\textit{(2) The second challenge stems from the unique~\mbox{MU-specific} strategic interactions between the auditor and~\mbox{the AI operator.}} 
The AI operator must balance model utility against compliance costs, since stricter MU will degrade model performance~\cite{sekhari2021remember}. This creates distinctive strategic tradeoffs absent in traditional auditing contexts. Existing audit frameworks (e.g., \cite{finley1994game, becker1968crime, tsebelis1990penalty,cho2019combating, plambeck2016supplier}), designed for conventional economic contexts, neither capture this MU-specific utility-compliance tradeoff nor accommodate probabilistic verification, rendering them inapplicable for MU auditing.

\textit{(3) The third major challenge lies in the intrinsic~analytical complexity induced by MU-specific nonlinearities.}~These~nonlinearities come from both the model utility and the detection probability that encodes interdependence between the auditor's inspections and the AI operator's compliance strategy, creating complex strategic couplings absent in traditional auditing (e.g., \cite{cho2019combating, plambeck2016supplier}). Such complexity precludes closed-form solutions, necessitating the development of new analytical techniques to establish equilibrium existence, uniqueness, and key structural properties.
\vspace{-10pt}
\subsection{Contributions}

In this paper, we introduce an economic auditing framework for enforcing MU compliance. Leveraging certified unlearning theory (e.g., \cite{guo2019certified,sekhari2021remember}) as the theoretical foundation, we model the MU-specific utility-compliance trade-off and characterize MU verification uncertainty through a hypothesis-testing interpretation to quantify auditor detection capability. We propose a simultaneous game to analyze the strategic interaction between the AI operator and the auditor under the prevalent undisclosed auditing practices, and extend the analysis to disclosed auditing through a sequential game to explore the impact of auditing transparency. We summarize our key contributions as follows.

\begin{itemize}
\item \textit{\mbox{New Problem Formulation.}}~To the best of our knowledge, we introduce the first economic framework for auditing MU compliance, integrating certified unlearning theory with regulatory enforcement to address the fundamental gap between MU's technical capabilities and regulatory implementation. Our framework addresses a critical need in emerging AI governance, as AI systems face increasing deletion requests under tightening privacy laws.
\item \textit{\mbox{Tractable Equilibrium Analysis.}} Our game analysis confronts significant analytical challenges from MU-induced nonlinearities in the model utility and detection probability, which create complex strategic couplings absent in the traditional auditing and preclude closed-form solutions. To address the challenge, we develop an auxiliary transformation technique that converts the complex bivariate nonlinear fixed-point problem into a tractable univariate formulation, which enables us to establish equilibrium existence, uniqueness, and structural properties. 


\item \textit{Counterintuitive Regulatory Insights.}
Our analysis reveals an unlearning audit paradox: the auditor could optimally reduce inspection intensity as deletion requests increase, since the AI operator's weakened unlearning makes non-compliance easier to detect. This aligns with recent audit reductions in China despite growing deletion volumes. 

\item \mbox{\textit{Auditing Transparency Paradox.}}
We prove that disclosed auditing consistently achieves superior cost-effectiveness compared to undisclosed auditing through a commitment mechanism. This then challenges the widespread reliance on undisclosed auditing practices in existing data governance.

\item \mbox{\textit{Performance Evaluation.}}
Experimental results using real data show that disclosed auditing increases the auditor's payoff by up to $2549.30\%$ and the AI operator's payoff by up to $74.60\%$ compared to the state-of-the-art benchmark. Moreover, the results also demonstrate that disclosed auditing outperforms undisclosed auditing for both players, leading to a win-win outcome through enhanced auditing transparency. 
\end{itemize}

The remainder of the paper is organized as follows. Section \ref{sec:related_work} reviews the related work, and Section~\ref{sec:system_model} presents the system model. Section~\ref{sec:equilibrium_analysis} analyzes the NE, and Section~\ref{sec:audit_implications} examines its regulatory implications. Section~\ref{sec:disclosure} further explores the impact of auditing transparency. Section~\ref{sec:experiment} presents simulations evaluating our framework based on real-data experiments, and Section~\ref{sec:conclusion} concludes the paper.

\section{Related Work}\label{sec:related_work}

\subsection{Foundations of MU}
MU, introduced by Cao and Yang in 2015 in \cite{cao2015towards},~has~spurred research on efficient methodologies for its technical implementations, which can be broadly categorized into exact unlearning (e.g., \cite{bourtoule2021machine}) and approximate unlearning (e.g., \cite{guo2019certified, sekhari2021remember, qiao2024hessian}). While exact unlearning through retraining guarantees perfect data removal, it is computationally prohibitive for large-scale models. Researchers have thereby shifted toward approximate methods that directly modify model parameters with reduced overhead. In particular, approximate unlearning can be certified when the distance between unlearned and retrained models is provably bounded \cite{guo2019certified,sekhari2021remember}, and this certified unlearning theory forms a key foundation of our research. Moreover, a few recent studies have examined the economic implications of MU (e.g., \cite{ding2025incentivized, wang2025unlearning}), focusing on interactions between users and AI operators through pricing and incentive mechanisms. 

However, these prior studies assumed AI operators voluntarily comply with deletion requests, which is unrealistic without external enforcement mechanisms. In practice, self-interested operators often deviate strategically \cite{a2021_california}, especially given the economic disincentives and information asymmetry inherent in MU. A primary goal of this research is to develop a theoretical analysis of AI operators' strategic non-compliance and propose a regulatory enforcement mechanism for MU compliance.

\subsection{MU Auditing and Verification Techniques}

In response to the potential risk of strategic non-compliance in MU, emerging research has developed technical approaches for auditing and verifying MU. These include backdoor-based methods (e.g., \cite{sommer2020towards,guo2023verifying}), reproduction-based strategies (e.g., \cite{thudi2022necessity,weng2024proof}), non-membership inference attacks (e.g., \cite{wang2024has}),~and model-difference techniques (e.g., \cite{wang2025evaluation}), among others (e.g., \cite{xu2024really,zhou2025truvrf}; see \cite{xue2025towards} for a survey). Although these technical approaches provide valuable tools for detecting non-compliance, they focus on implementation challenges rather than regulatory design. A key overlooked perspective is the economic dimension necessary for effective enforcement: incorporating penalty schemes to incentivize compliance and designing MU auditing strategies that balance enforcement costs with effectiveness. This paper addresses this gap by introducing the first economic framework for auditing MU compliance.

\subsection{Traditional Economic Auditing Framework}

Economic auditing has been extensively studied in financial accounting (e.g., \cite{finley1994game}), law enforcement (e.g., \cite{becker1968crime, tsebelis1990penalty}), and supply chain management (e.g., \cite{cho2019combating, plambeck2016supplier}). These traditional frameworks typically model auditing as a strategic interaction between the auditor and the auditee, mainly focusing on how inspections and penalties influence compliance. However, existing frameworks are inadequate for auditing MU compliance due to three fundamental challenges. First, MU introduces inherent uncertainty through the statistical guarantees of certified unlearning theory~\cite{guo2019certified,sekhari2021remember}, which is fundamentally different from the traditional data deletion auditing that allows for deterministic verification. Second, existing models lack a theoretical foundation to quantify MU deletion effectiveness and overlook the compliance-utility tradeoff, wherein stricter compliance reduces model performance. Third, existing auditing mechanism design, including penalty schemes and detection structures, are either too generic or tailored to specific traditional contexts, making them unsuitable for MU. 

To the best of our knowledge, this paper introduces the first economic framework for auditing MU compliance. A key novelty is integrating certified unlearning with regulatory auditing to theoretically model MU's inherent probabilistic verification and the utility-compliance tradeoff. On the technical level, we address MU-induced nonlinearities that create complex strategic couplings absent in traditional auditing through auxiliary transformations and analytically derive equilibrium properties that reveal regulatory implications for MU auditing.

\section{System Model}\label{sec:system_model}

In this section, we introduce a game-theoretic framework for MU compliance auditing. This framework captures the strategic interaction between an AI operator, who will determine the unlearning certification level by trading off utility loss against audit risk, and an auditor, who selects the inspection intensity to balance detection effectiveness against enforcement costs.\footnote{For the sake of clarity, we will henceforth refer to the auditor using female pronouns and the AI operator using male pronouns.}

In what follows, Section~\ref{subsec:MU_compliance} outlines our proposed framework, which is based on the certified unlearning theory as the theoretical foundation of our model. Section~\ref{subsec:AI_Operator} then elaborates on the model for the AI operator's strategic MU decision, while Section~\ref{subsec:model_auditor} develops the auditor's compliance auditing model. Finally, Section~\ref{sec:game_formulation} formulates the MU compliance auditing game.

\subsection{Framework for MU Compliance Auditing}\label{subsec:MU_compliance}

To illustrate our framework, we introduce a real-world motivating scenario, as depicted in Fig.~\ref{fig:illustration_framework}. Consider the interaction between an AI operator (e.g., Google) and an auditor (e.g., the DPA) in the context of MU compliance. When users request the deletion of their personal data (e.g., search history) from a trained model (e.g., Google Autocomplete for query predictions), the AI operator employs MU techniques to remove data influence. Meanwhile, the auditor conducts compliance audits to verify data deletion effectiveness and imposes penalties for detected non-compliance. Since exact unlearning via complete retraining is infeasible for large-scale models in practice, we assume the operator employs approximate unlearning methods \cite{guo2019certified,sekhari2021remember,qiao2024hessian}.

Our framework is grounded in the \emph{($\epsilon,\delta$)-certified unlearning theory} \cite{sekhari2021remember}, which measures the effectiveness of approximate unlearning methods, by providing theoretical guarantees that an unlearned model is statistically indistinguishable from one trained without the deleted data, as formally defined below:

\begin{definition}[$(\epsilon,\delta)$-certified unlearning \cite{sekhari2021remember}]\label{def:certified_unlearning}
For any original dataset $\mathcal{D}$ and forgetting dataset $\mathcal{D}_f \subseteq \mathcal{D}$, and for any hypothesis space subset $H \subseteq \mathcal{H}$, an unlearning algorithm $\mathcal{R}$ is $(\epsilon,\delta)$-unlearning for a learning algorithm $\mathcal{A}$ if:
\begin{equation*}
\text{\emph{Prob}}\left(\mathcal{R}\left(\mathcal{A}(\mathcal{D})\right)\in H\right)\le \exp(\epsilon)\cdot\text{\emph{Prob}}\left(\mathcal{A}(\mathcal{D}\setminus \mathcal{D}_f)\in H\right)+\delta,
\end{equation*}
and
\begin{equation*}
\text{\emph{Prob}}\left(\mathcal{A}(\mathcal{D}\setminus \mathcal{D}_f)\in H\right)\le \exp(\epsilon)\cdot\text{\emph{Prob}}\left(\mathcal{R}\left(\mathcal{A}(\mathcal{D})\right)\in H\right) +\delta,
\end{equation*}
where $\epsilon>0$ is the unlearning certification level, and $\delta\in[0,1]$ is the privacy failure parameter.
\end{definition}

The \textit{unlearning certification level} $\epsilon$ measures data deletion effectiveness: a smaller $\epsilon$ indicates more thorough data removal (stricter unlearning) but greater model utility loss, as quantified in Section~\ref{subsec:AI_Operator}. Therefore, this creates the fundamental trade-off underlying our compliance auditing game, wherein the operator balances user privacy protection (and thus compliance) against model performance under auditing uncertainty. Notice that $\epsilon$ is not directly observable by the auditor and can only be evaluated through probabilistic verification (Section \ref{subsec:model_auditor}).

\subsection{AI Operator's Strategic MU Decision}\label{subsec:AI_Operator}

This section models the AI operator's strategic MU decision grounded in the certified unlearning theory, and discusses the resulting model utility and underlying trade-off.

\subsubsection{MU Decision}\label{subsubsec:mu_decision}
The operator's strategic decision involves selecting an unlearning certification level $\epsilon \in [0,\infty]$ to balance model utility with audit risk. Here, $\epsilon= \infty$ represents complete disregard for deletion requests, smaller $\epsilon$ values imply stricter unlearning and thereby better MU compliance, and the extreme $\epsilon= 0$ corresponds to complete data removal.

\subsubsection{Model Utility Function}\label{subsubsec:model_utility_function} 

Following the existing literature on economic implications of AI models (e.g., \cite{donahue2021model}), we assume that the AI operator's utility from performing MU depends on the unlearned model's generalization performance. Since MU typically degrades performance, we define the model utility function as the negative test loss of the unlearned model, i.e., $u(\epsilon) < 0$, which captures the performance degradation induced by MU.\footnote{We consider the AI operator's model utility in the no-unlearning baseline to be a constant. Mathematically, it does not affect our analysis, so we directly normalize the model utility as the negative test loss in this paper.} Based on the theoretical upper bound of test loss in \cite{sekhari2021remember}, we have:
\begin{equation}\label{equ:model_loss_simplify}
    u(\epsilon)=-\frac{G\eta^2}{\epsilon}.
\end{equation}
where the parameter $G\triangleq\sqrt{d} M L^3\sqrt{\ln\left(1/\delta\right)}/{\lambda^3}$ consolidates model characteristics: the model dimension $d$, Lipschitz constant~$L$, strong convexity parameter~$\lambda$, and Hessian Lipschitz constant $M$. The parameter $\eta \triangleq \omega/n$  represents the unlearning data proportion~(deleted samples $\omega$ over total samples $n$), with $\eta\in[0,\eta_{\max}]$ wherein $\eta_{\max}\le~1$. 

Although derived from~\cite{sekhari2021remember}, the model utility function~in~\eqref{equ:model_loss_simplify} reflects a common pattern in the theoretical guarantees of~MU algorithms \cite{suriyakumar2022algorithms,liu2023certified}: the test loss upper bound typically scales inversely with~$\epsilon$ and quadratically with $\eta$. Moreover, we also empirically validate the inverse relationship between $u(\epsilon)$ and $\epsilon$ predicted by \eqref{equ:model_loss_simplify} in Section~\ref{subsubsec:frame_validation} through real-data MU experiments.

\subsubsection{Strategic Trade-off}\label{subsubsec:model_utility_discussions} 
The model utility function~$u(\epsilon)$~in \eqref{equ:model_loss_simplify} increases concavely with the unlearning certification level $\epsilon$ and decreases concavely with the unlearning data proportion~$\eta$. This captures a key MU feature: stricter unlearning (smaller $\epsilon$) reduces model utility $u(\epsilon)$, with a sharper decrease at higher $\eta$. However, a higher $\epsilon$ mitigates model utility loss but increases audit risk, creating the central strategic tension in our game.

\subsection{Auditor's Compliance Auditing Model}\label{subsec:model_auditor}

\begin{figure}[t]
	\centering  
	\includegraphics[width=\linewidth]{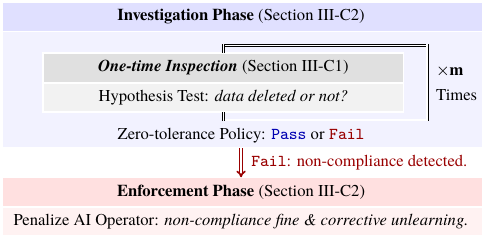}
    \vspace{-15pt}
	\caption{Auditor's Two-stage Compliance Auditing Process: Investigation and Enforcement. In the investigation phase, the auditor performs $m$ independent inspection tests via probabilistic MU verification and declares non-compliance if any test fails (zero-tolerance). In the enforcement phase, the auditor imposes a fine and mandates corrective unlearning upon detecting non-compliance.}
	\label{fig:Deletion_Compliance_Model}
    \vspace{-10pt}
\end{figure}

The auditor's compliance auditing follows a two-stage process: an investigation phase to detect non-compliance, and an enforcement phase to impose penalties, as illustrated in Fig.~\ref{fig:Deletion_Compliance_Model}.

To quantify the inherent uncertainty of MU verification, we introduce a hypothesis-testing interpretation of $(\epsilon,\delta)$-certified unlearning in Section \ref{subsubsec:Hypothesis_unlearning}, which helps formalize the auditor's non-compliance detection capability. Building upon this, we formalize the two-stage auditing process in Section \ref{subsubsec:interim_audit}, specifying both the investigation phase and the enforcement mechanisms.

\subsubsection{\mbox{Hypothesis-Testing Analysis of MU Verification}}\label{subsubsec:Hypothesis_unlearning} 
To begin, we model each independent auditor inspection (conducted via MU verification) as a hypothesis test. Next, we introduce a hypothesis-testing framework grounded in the $(\epsilon,\delta)$-certified unlearning, and then derive the auditor's detection capability.

\textbf{Hypothesis-testing Framework.} The auditor tests whether the target dataset $\mathcal{D}_f$ has been effectively deleted by comparing the observed model against two competing hypotheses: (i) \emph{Null}: the model is trained on $\mathcal{D} \setminus \mathcal{D}_f$; (ii) \emph{Alternative}: the model is trained on $\mathcal{D}$ and then unlearned. The null indicates successful deletion of $\mathcal{D}_f$ from the model trained on $\mathcal{D}$, while the alternative indicates residual dependence on $\mathcal{D}_f$. 

In the following lemma, we formalize the hypothesis-testing interpretation of $(\epsilon,\delta)$-certified unlearning, drawing inspiration from the proof techniques for $(\epsilon,\delta)$-differential privacy in \cite{kairouz2015composition}.

\begin{lemma}[Hypothesis-testing interpretation of $(\epsilon,\delta)$-certified unlearning, inspired by \cite{kairouz2015composition}]\label{The:hypothesis}
For any~$\epsilon>0$ and $\delta\in[0,1]$, an unlearning algorithm $\mathcal{R}$ is $(\epsilon,\delta)$-unlearning for a learning algorithm $\mathcal{A}$ if and only if the following conditions are satisfied for any original dataset $\mathcal{D}$ and forgetting dataset $\mathcal{D}_f\subseteq\mathcal{D}$, and all rejection region $S\subseteq\mathcal{H}$:
\begin{equation*}
\emph{\text{PFA}}+\exp(\epsilon)\cdot\emph{\text{PMD}}\ge 1-\delta,\text{ and }\exp(\epsilon)\cdot\emph{\text{PFA}}+\emph{\text{PMD}}\ge 1-\delta,
\end{equation*}
wherein $\text{\emph{PFA}}$ and $\text{\emph{PMD}}$ are false alarm and missed detection probabilities.
\end{lemma}

\textbf{Auditor's Detection.} Given the inherent uncertainty of MU verification, we next quantify the auditor's detection capability in each inspection test. Specifically, we derive the probability that the auditor fails to detect non-deletion by the AI operator, corresponding to the PMD in the hypothesis-testing framework (Lemma~\ref{The:hypothesis}), as formalized below.

\begin{lemma}[MU verification uncertainty]\label{lemma:error}
For $(\epsilon,\delta)$-certified unlearning under practical parameters, the probability that the auditor fails to detect non-deletion in a single inspection, as characterized by the $\text{\emph{PMD}}$ in the hypothesis testing, is lower-bounded by
\begin{equation}\label{equ:error_rate}
    \xi\cdot \exp(-\epsilon),
\end{equation}
where $\xi \triangleq1-\delta-\text{PFA}$ is a composite parameter capturing deletion failure and false alarm rate, typically close to one.\footnote{In practice, $\delta$ is typically cryptographically small (e.g., $10^{-9}$), while $\text{PFA}$ is commonly set to $0.05$ or lower in statistical hypothesis testing.}
\end{lemma}

\subsubsection{Two-stage Compliance Auditing Model}\label{subsubsec:interim_audit}

We next model the auditor's two-stage compliance auditing process, as illustrated in Fig.~\ref{fig:Deletion_Compliance_Model}.

\mbox{\textbf{Auditing Process.}} In the investigation phase, the auditor chooses an \textit{inspection intensity} $m \geq 0$, which represents the number of independent inspection tests.\footnote{For simplicity, we allow non-integer $m$, with the auditor performing audits in a randomized manner to ensure the expected number of inspections equals $m$. This approach is common in modeling targeting and sampling processes, wherein it has minimal impact on welfare outcomes \cite{yu2017public}.}\textsuperscript{,}\footnote{For tractability, we assume independent inspections, a common assumption in modeling targeting and sampling processes (e.g.,~\cite{yu2017public}). In practice, however, MU verification tests applied to the same model may exhibit correlated errors, potentially reducing the marginal benefit of additional inspections. Examining correlated inspections is an important direction for future work.} Particularly, a larger value of $m$ indicates more intensive auditing or greater auditor effort. We assume that the auditing process incurs a linear cost of $mc$, where $c$ is the unit cost per inspection, for tractability.

Moreover, we assume that the AI operator is deemed compliant if and only if all inspections are successfully passed; any failure results in non-compliance. This assumption reflects the \textit{zero-tolerance policy} for regulatory auditing, which is widely adopted in practice (e.g., by the EU DPA \cite{schwartz_2021_gdpr}) to ensure strict compliance and mitigate risk.

Applying this zero-tolerance policy, and based on Lemma~\ref{lemma:error}, the auditor's probability of detecting non-compliance is given by\footnote{For tractability, we assume the auditor can ideally adopt the most powerful MU verification test, achieving a per-inspection $\text{PMD}$ of $\xi\cdot \exp(-\epsilon)$, the lower bound from Lemma~\ref{lemma:error}. This bound-based approach is standard in developing economic frameworks for statistical learning (e.g., \cite{ding2025incentivized}).}
\begin{equation}\label{equ:Audit_accuracy}
\text{Detect}(m,\epsilon)\triangleq1-\xi^m\exp\left(-m\epsilon\right), 
\end{equation}
referred to as the \textit{detection probability}, which increases with and is concave in both $m$ and $\epsilon$.

\mbox{\textbf{Enforcement Mechanism.}} In the enforcement phase, upon detecting non-compliance, the auditor will impose two types of penalties following real-world practices such as the GDPR~\cite{gdpr_2018_general}: (i) \textit{a non-compliance fine} $p$, which serves as tax revenue \cite{fainmesser2023digital}, and (ii) \textit{mandatory corrective unlearning}, which mandates the operator to achieve at least the minimum deletion level $\epsilon_0>0$, causing a model utility loss of $u(\epsilon)-u(\epsilon_0)$ per \eqref{equ:model_loss_simplify}. Together, these penalties create audit risk that constrains the operator's strategic MU decision (as modeled in Section \ref{subsec:AI_Operator}).

\subsection{Game Formulation}\label{sec:game_formulation}

Finally, we formally formulate the operator-auditor interaction as a simultaneous game, where neither player observes the other's action before choosing their own decision. This reflects prevalent real-world auditing practices, such as unannounced or undisclosed inspections conducted by EU and French DPAs \cite{gobert_2024_audit,a2022_practical}. We next introduce the players' payoffs and information structure, and conclude with the formal game definition.

\textbf{Player Payoffs.} Due to the probabilistic nature of auditing \cite{sommer2020towards}, we now formulate the expected payoffs for both players considering all possible audit outcomes:
\begin{itemize}
\item \emph{Auditor's Payoff}: The auditor is both socially responsible and welfare-oriented, valuing the MU compliance level as well as the regulatory surplus (e.g., for future regulatory budgets). We define the regulatory surplus as the expected fine revenue minus the auditing costs:
\begin{equation}\label{Equ:payoff_auditor_surplus}
    \Pi_c(m;\epsilon)\triangleq p \cdot \text{Detect}(m,\epsilon) - mc,
\end{equation}
where a higher auditing intensity $m$ increases the detection probability $\text{Detect}(m,\epsilon)$ in \eqref{equ:Audit_accuracy}, but also raises costs. The auditor's overall payoff is then given by
\begin{equation}\label{Equ:payoff_auditor}
\Pi(m;\epsilon) \triangleq \gamma_1\Pi_c(m;\epsilon)+\gamma_2(-\epsilon),
\end{equation}
where $\Pi_c(m;\epsilon)$ denotes the regulatory surplus defined in \eqref{Equ:payoff_auditor_surplus}. The term $-\epsilon$ reflects the social concern for improved MU compliance, indicating that a lower certification level $\epsilon$ is desirable. Parameters $\gamma_1>0$ and $\gamma_2>0$ represent the relative weights on financial returns and social concerns, respectively. The auditor's payoff in \eqref{Equ:payoff_auditor} hence reflects the cost-effectiveness of regulatory enforcement.

\item \emph{AI Operator's Payoff}: The operator's payoff equals the model utility in \eqref{equ:model_loss_simplify} minus expected non-compliance costs, which consist of the non-compliance fine $p$ and the model utility loss resulting from corrective unlearning:
\begin{equation}\label{Equ:payoff_operator}
\pi(\epsilon;m) \triangleq u(\epsilon) - \left(p + u(\epsilon) - u(\epsilon_0)\right) \cdot \text{Detect}(m,\epsilon),
\end{equation}
balancing model performance degradation from unlearning against audit failure risk.
\end{itemize}

\mbox{\textbf{Information Structure.}} We consider complete information where the auditor knows the model characteristic parameter $G$ (e.g., through AI model registries under the EU AI Act~\cite{a2016_article}) and unlearning data proportion $\eta$ (e.g., via GDPR investigative powers~\cite{gdpr_art}). However, as neither player observes the other's action, mirroring the unannounced and undisclosed audits that are common in practice\cite{gobert_2024_audit,a2022_practical}, the interaction is modeled as a simultaneous game with imperfect information.

We formally define this game as follows.

\begin{definition}[MU Compliance Auditing Game] The MU~compliance auditing game consists of:\footnote{Throughout this study, we treat the non-compliance fine~$p$ as an exogenous parameter, reflecting real-world settings where penalty decisions are typically set by external authorities. In Section \ref{subsubsec:fine}, we use simulations to explore how $p$ affects MU compliance and offer preliminary insights into setting the fine to optimize social costs.}
\begin{itemize}
    \item \emph{Players:} The AI operator and the auditor.
    \item \emph{Strategies:} The AI operator chooses the unlearning certification level $\epsilon$ from the strategy set $[0, \infty]$, while the auditor decides the inspection intensity $m$ based on the strategy set $[0,\infty)$.
    \item \emph{Payoffs:} The AI operator maximizes his expected payoff $\pi(\epsilon,m)$ in \eqref{Equ:payoff_operator}, while the auditor maximizes her expected payoff $\Pi(m,\epsilon)$ in \eqref{Equ:payoff_auditor}.
\end{itemize}
\end{definition}  

The solution concept of our interest is the \textit{Nash equilibrium (NE)}, wherein no player has an incentive to deviate unilaterally \cite{fudenberg1991game}. Different from traditional auditing frameworks (e.g., \cite{cho2019combating, plambeck2016supplier}), our model introduces MU-specific nonlinearities arising from both the model utility function in \eqref{equ:model_loss_simplify} and the detection probability in \eqref{equ:Audit_accuracy}. These nonlinearities generate intricate strategic couplings that are absent in classic auditing games and preclude closed-form NE solutions. In Section~\ref{subsec:BR_auditor}, we further elaborate on these technical challenges and show how we address them analytically through auxiliary transformations.

For clarity, we assume $\xi=1$ throughout our main analysis, where $\xi$ is the composite parameter capturing deletion failure and false alarm rate, typically close to one in practice, as stated in Lemma \ref{lemma:error}. Appendix C relaxes this assumption and confirms our main insights remain robust under general $\xi$.

\section{Equilibrium Analysis}\label{sec:equilibrium_analysis}

This section analyzes the NE of our MU compliance auditing game. Specifically, we first characterize both players' optimal strategies through best-response analysis (Section~\ref{subsec:BR_deletion}), and then establish the existence and uniqueness of equilibrium (Section~\ref{sec:NE_analysis}).

\subsection{Best Response Analysis}\label{subsec:BR_deletion}

In this subsection, we perform best-response analysis~for~the AI operator and the auditor to characterize their optimal~strategies, respectively.

\subsubsection{AI Operator's Optimal Unlearning Certification Level}\label{subsubsec:AI}
We first analyze the AI operator's optimal certification level $\epsilon^*(m)$, given the auditor's inspection intensity $m$. 

Recall that the operator chooses $\epsilon^*$ to maximize his expected payoff $\pi(\epsilon;m)$ in \eqref{Equ:payoff_operator}. While this payoff function in~\eqref{Equ:payoff_operator}~is~generally non-convex in $\epsilon$, its unimodal structure enables a closed-form characterization of $\epsilon^*(m)$, as established in the following proposition.

\begin{proposition}\label{Prop:BR_certification}
Given the auditor's inspection intensity $m$, the AI operator's optimal unlearning certification level is given by
\begin{equation}\label{Equ:BR_certification}
\epsilon^*(m)=\frac{mG\eta^2\epsilon_0+\sqrt{m^2G^2\eta^4\epsilon_0^2+4G\eta^2m\left(G\eta^2\epsilon_0+p\epsilon_0^2\right)}}{2m\left(G\eta^2+p\epsilon_0\right)}
\end{equation}
when $m> 0$; otherwise, when $m=0$, $\epsilon^*(m)=\infty$. Moreover, $\epsilon^*(m)$ decreases convexly with $m$.
\end{proposition}

Proposition \ref{Prop:BR_certification} suggests that any positive inspection intensity ($m \neq 0$) ensures a finite certification level $\epsilon^*(m)<\infty$~in~\eqref{Equ:BR_certification}, preventing the operator from ignoring deletion requests. As $m$ increases, the operator reduces $\epsilon^*(m)$ to lower audit risk, confirming that stricter auditing incentivizes better compliance.

However, the marginal effectiveness of intensified auditing diminishes, as suggested by the convexity of $\epsilon^*(m)$ in $m$. This diminishing return stems from a key characteristic of MU (see Section \ref{subsec:AI_Operator}): the concave decline of model utility $u(\epsilon)$ in \eqref{equ:model_loss_simplify} as $\epsilon$ decreases, making further reductions in $\epsilon$~(and thus stricter compliance) prohibitively costly, even under intensified audits.

We next examine how the non-compliance fine~$p$ shapes the operator's compliance strategy $\epsilon^*(m)$.

\begin{corollary}\label{Cor:BR_certification_penalty}
Given the auditor's inspection intensity $m$, the AI operator's optimal unlearning certification level $\epsilon^*(m;p)$ decreases convexly in the non-compliance fine $p$.
\end{corollary}


Intuitively, a higher fine $p$ increases the cost of audit failure, prompting the operator to lower $\epsilon$ to lower audit risk. However, the marginal impact diminishes as $p$ increases, as indicated by the convexity of $\epsilon^*(m; p)$ in $p$. The explanation here parallels the reasoning for Proposition~\ref{Prop:BR_certification}. We refer to this phenomenon as the \textit{penalty saturation effect}, which persists in equilibrium, as shown by the numerical results in Section~\ref{subsubsec:fine}. To avoid repetition, we defer further discussions on policy insight there. 

These above findings lead to our first key observation:

\begin{observation}\label{Obs:BR_marginal}
Both audits and fines exhibit diminishing marginal effectiveness in regulating the AI operator's unlearning certification level as these measures intensify.
\end{observation}

\subsubsection{The Auditor's Optimal Inspection Intensity}\label{subsec:BR_auditor} 

Now, we analyze the auditor's optimal inspection intensity $m^*(\epsilon)$, given any unlearning certification level $\epsilon$ chosen by the AI operator. 

Recall that the auditor chooses an inspection intensity $m^*$ to maximize her expected payoff $\Pi(m;\epsilon)$ in \eqref{Equ:payoff_auditor}, given the operator's certification level $\epsilon$. In principle, this involves~balancing two objectives: maximizing the regulatory surplus $\Pi_c(m;\epsilon)$ in \eqref{Equ:payoff_auditor_surplus} against minimizing the social concern tied to the operator's certification level $\epsilon$. However, the undisclosed nature of inspection intensity means it cannot directly influence the operator's strategic choice of $\epsilon$. Consequently, the auditor~focuses solely on maximizing the regulatory surplus $\Pi_c(m;\epsilon)$ in \eqref{Equ:payoff_auditor_surplus}. Since $\Pi_c$ is concave in $m$, we derive a closed-form characterization of $m^*(\epsilon)$ in the following proposition.

\begin{proposition}\label{Prop:BR_audit}
Given the AI operator's unlearning certification level $\epsilon$, the auditor's optimal inspection intensity $m^*(\epsilon)$ is~given~by
\begin{equation}\label{Equ:BR_audit}
    m^*(\epsilon)=\left\{
    \begin{aligned}
        &\frac{1}{\epsilon}\ln\left(\frac{p\epsilon}{c}\right),&\text{if}\ \ \epsilon>\frac{c}{p},\\
        &0, &\text{otherwise}.
    \end{aligned}
    \right.
\end{equation}
\end{proposition}


Proposition \ref{Prop:BR_audit} indicates that the optimal inspection intensity $m^*(\epsilon)$ decreases with auditing cost $c$ and increases with~fine~$p$. For high cost-to-fine ratios (i.e., $ c/p\geq \epsilon$), for instance, under weak regulatory infrastructure or complex MU verification, the auditor forgoes auditing (i.e., $m^*=0$), since the expected fine revenue cannot cover auditing costs.

\begin{figure}[t]
    \centering
    \vspace{-10pt}
    \subfigure[$m^*(\epsilon)$ versus $\epsilon$.]{
        \begin{minipage}{0.22\textwidth}
            \centering
            \begin{tikzpicture}[scale=1]
            \draw [->,>=latex] (-0.25,0) -- (3,0) node [below]  {\small $\epsilon$};
            \draw [->,>=latex] (0,-0.25) -- (0,1.5) node [right]  {\small $m^*(\epsilon)$};
            \draw [thick,color=red] (0,0) -- (0.4,0);
            \draw [thick,color=red,domain=0.4:2.7,samples=200] plot (\x, {1/\x*ln(5/2*\x)});
            \draw[color=red,dotted] (0.4*2.7182,0) -- (0.4*2.7182,1/0.4/2.7182);
            \draw[color=red,thick] (0.4*2.7182,-0.05) -- (0.4*2.7182,0.05);
            \node at (0,0) [below left] {\small $0$};
            \node at (0.4,0) [color=red,below] {\small $\hat{\epsilon}$};
            \node at (0.4*2.7182,0) [color=red,below] {\small $\tilde{\epsilon}$};
            \fill[red] (0.4,0) circle (1pt);
            \fill[red] (0.4*2.7182,1/0.4/2.7182) circle (1pt);
            \end{tikzpicture}
            \label{fig:Auditor_response_intensity}
		\end{minipage}
	}
    \centering
    \subfigure[$\text{Detect}\left(m^*(\epsilon),\epsilon\right)$ versus~$\epsilon$.]{
        \begin{minipage}{0.22\textwidth}
            \centering
            \begin{tikzpicture}[scale=1]
            \draw [->,>=latex] (-0.25,0) -- (3,0) node [below]  {\small $\epsilon$};
            \draw [->,>=latex] (0,-0.25) -- (0,1.5) node [right]  {\small $\text{Detect}\left(m^*(\epsilon),\epsilon\right)$};
            \draw [thick,color=blue] (0,0) -- (0.4,0);
            \draw [thick,color=blue,domain=0.4:2.7,samples=200] plot (\x, {1-0.4/\x});
            \draw[color=blue,dotted] (0.4*2.7182,0) -- (0.4*2.7182,1-1/2.7182);
            \draw[color=blue,thick] (0.4*2.7182,-0.05) -- (0.4*2.7182,0.05);
            \node at (0,0) [below left] {\small $0$};
            \node at (0.4,0) [color=blue,below] {\small $\hat{\epsilon}$};
            \node at (0.4*2.7182,0) [color=blue,below] {\small $\tilde{\epsilon}$};
            \fill[blue] (0.4,0) circle (1pt);
            \fill[blue] (0.4*2.7182,1-1/2.7182) circle (1pt);
            \end{tikzpicture}
        	\label{fig:Auditor_response_accuracy}
    	\end{minipage}
	}
    \vspace{-5pt}
    \caption{The auditor's optimal inspection intensity $m^*(\epsilon)$ in \eqref{Equ:BR_audit} and detection probability $\text{Detect}\left(m^*(\epsilon),\epsilon\right)$ in \eqref{equ:Audit_accuracy} versus unlearning certification level $\epsilon$.}
	\label{fig:Auditor_response}
    \vspace{-10pt}
\end{figure}
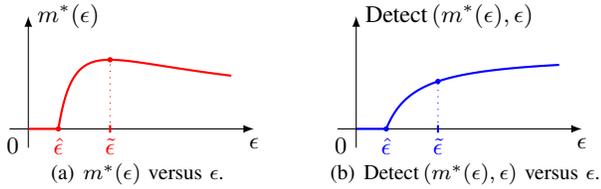

To evaluate how the auditor responds to varying certification levels $\epsilon$, the following corollary (together with Fig.~\ref{fig:Auditor_response}) analyzes their impact on the optimal inspection intensity $m^*(\epsilon)$~in~\eqref{Equ:BR_audit} and corresponding detection probability in \eqref{equ:Audit_accuracy}. For clarity, we define two \textit{unlearning certification thresholds}:
\begin{equation}\label{equ:deletion_threshold}
\hat{\epsilon}\triangleq c/p\quad\text{and}\quad\tilde{\epsilon}\triangleq ec/p,
\end{equation}
where $e$ is the base of the natural logarithm and $\hat{\epsilon}<\tilde{\epsilon}$. 

\begin{corollary}\label{Cor:BR_Audit}
The impact of the AI operator's unlearning~certification level $\epsilon$ is summarized as follows (see Fig. \ref{fig:Auditor_response}).
\begin{itemize}
    \item [(i)] The optimal inspection intensity $m^*(\epsilon)$ remains zero for $\epsilon\in(0,\hat{\epsilon})$, increases over $[\hat{\epsilon},\tilde{\epsilon})$, and decreases over $[\tilde{\epsilon},\infty)$ (see Fig. \ref{fig:Auditor_response_intensity}).
    \item [(ii)] The detection probability $\text{Detect}\left(m^*(\epsilon),\epsilon\right)$ remains zero for $\epsilon\in(0,\hat{\epsilon})$ and increases monotonically for $[\hat{\epsilon},\infty)$ (see Fig. \ref{fig:Auditor_response_accuracy}).
\end{itemize}
\end{corollary}

Corollary \ref{Cor:BR_Audit} reveals a non-monotonic pattern in the auditor's optimal inspection intensity $m^*(\epsilon)$ as $\epsilon$ varies, as shown in Fig. \ref{fig:Auditor_response_intensity}. Specifically, for low levels $\epsilon\in(0,\hat{\epsilon})$, the auditor refrains from auditing because detecting non-compliance is less likely. As $\epsilon$ rises into $[\hat{\epsilon},\tilde{\epsilon})$, detection becomes more likely, prompting the auditor to intensify inspections $m$. As a result, the detection probability of $\text{Detect}\left(m^*(\epsilon),\epsilon\right)$ rises, as depicted in Fig. \ref{fig:Auditor_response_accuracy}.

However, once the certification level $\epsilon$ exceeds the threshold $\tilde{\epsilon}$, non-compliance becomes readily detectable, and further increases in inspection intensity yield diminishing returns.~Thus, the auditor lowers inspection intensity $m^*$ to cut auditing costs, as depicted in Fig. \ref{fig:Auditor_response_intensity}. Nonetheless, the detection probability $\text{Detect}(m^*(\epsilon), \epsilon)$ in Fig. \ref{fig:Auditor_response_accuracy} continues to increase, uncovering more non-compliance despite reduced inspection intensity. 

\subsection{Nash Equilibrium}\label{sec:NE_analysis}

Building upon the previous best-response analyses, we now derive the NE of our game. A standard approach is to combine the AI operator's optimal certification level $\epsilon^*(m)$ in~\eqref{Equ:BR_certification} (see Proposition~\ref{Prop:BR_certification}) with the auditor's optimal inspection intensity $m^*(\epsilon)$ in \eqref{Equ:BR_audit} (see Proposition~\ref{Prop:BR_audit}), and then identify their fixed point. This fixed point characterizes the NE of the game~\cite{fudenberg1991game}.

However, although the best-response functions in \eqref{Equ:BR_audit} and~\eqref{Equ:BR_certification} are well-established, directly solving the associated fixed-point problem remains challenging due to the nonlinearities induced by MU. These nonlinearities stem from the model utility in \eqref{equ:model_loss_simplify} and the detection probability in \eqref{equ:Audit_accuracy}, which jointly create complex strategic couplings that are absent in traditional auditing games. From a technical perspective, the resulting non-convex operator payoff gives rise to a complicated best-response function in \eqref{Equ:BR_certification}, involving nonlinear interactions among multiple parameters/variables. Meanwhile, the auditor's best response in \eqref{Equ:BR_audit} exhibits a piecewise and non-monotonic structure (see Corollary~\ref{Cor:BR_Audit}). These complexities collectively preclude the derivation of closed-form equilibrium solutions.

We address the above challenges by transforming the original bivariate fixed-point problem into a tractable univariate auxiliary problem, thereby decoupling the system. Specifically, we establish the existence and uniqueness of the equilibrium without relying on explicit closed-form solutions through three key steps (see Appendix~H for more technical details). First, we derive the necessary condition that any equilibrium solution must satisfy. Second, we leverage the special structure of our problem to reduce it to a univariate equation $\mathcal{J}(\epsilon)=0$, where
\begin{equation}\label{HHH}
        \mathcal{J}(\epsilon)\triangleq \text{Detect}\left(m^*(\epsilon), \epsilon\right)-\text{Detect}\left(m, \epsilon^*(m)\right),
\end{equation}
which compare detection probabilities under each player's best response. Finally, we establish the monotonicity of $\mathcal{J}(\epsilon)$, which implies a unique solution $\epsilon^*$ and hence a unique NE.

The following theorem formally establishes the existence and uniqueness of our NE.

\begin{theorem}\label{Theorem:NE}
There exists a unique NE as follows. Specifically, the AI operator's equilibrium unlearning certification level $\epsilon^*$ is the unique solution to
\begin{equation}\label{Equ:NE_certification}
    \left(\left(p+\frac{G\eta^2}{\epsilon_0}\right)\epsilon^*-G\eta^2\right)\ln\left(\frac{p \epsilon^*}{c}\right)=G\eta^2,
\end{equation}
and the auditor chooses a non-zero inspection intensity of
\begin{equation}\label{Equ:NE_audit}
    m^*=\frac{1}{\epsilon^*}\ln\left(\frac{p\epsilon^*}{c}\right).
\end{equation}
\end{theorem}

Theorem~\ref{Theorem:NE} reveals that the auditor maintains strictly positive inspection intensity in the equilibrium (i.e., $m^* > 0$), thereby preventing the AI operator from entirely disregarding deletion requests, as reflected by the finite certification level $\epsilon^*$~in~\eqref{Equ:NE_certification}. However, further equilibrium characterization is fundamentally constrained by the absence of a closed-form solution and the MU-specific strategic couplings. The next section therefore examines the equilibrium properties in greater details to uncover additional regulatory insights.

\section{Regulatory Implications}
\label{sec:audit_implications}

Building on the established NE in Theorem~\ref{Theorem:NE}, we now~investigate how the unlearning data proportion $\eta$ shapes equilibrium behaviors and derive key regulatory implications. Our analysis is guided by a central question: As AI systems encounter~rising deletion requests under increasingly tightened global privacy regulations, how should the auditor optimally adjust inspection intensity? Is intensifying inspections to safeguard data privacy always optimal, as the conventional wisdom suggests?

Specifically, Section~\ref{subsec:theoretical} first theoretically establishes how the auditor adjusts equilibrium inspection intensity in response to changes in~$\eta$. Section~\ref{subsec:comparative} compares the regulatory environments and auditing practices in the EU and China, demonstrating qualitative consistency with our theoretical predictions.

\subsection{Theoretical Analysis}\label{subsec:theoretical}

In this subsection, we theoretically analyze how the auditor adjusts equilibrium inspection intensity in response to changes in~$\eta$. However, this analysis of equilibrium properties remains complicated by the nonlinearities inherent in MU, which~create complex strategic couplings and preclude a closed-form NE characterization, making direct characterization challenging. 

To address these challenges, we again rely on the auxiliary function $\mathcal{J}(\epsilon)$ defined in \eqref{HHH} from the transformed equivalent problem introduced in Section~\ref{sec:NE_analysis}. Our analysis proceeds in three steps. To begin, we establish the relationship between the equilibrium unlearning certification level $\epsilon^*$ and the unlearning data proportion $\eta$ (Lemma~\ref{Prop:epsilon_deletion}). Next, we exploit the structural property of the auditor's non-monotonic best response to varying unlearning certification levels, which identifies a critical threshold $\tilde{\epsilon}$ in \eqref{equ:deletion_threshold} (Corollary~\ref{Cor:BR_Audit}). Finally, by linking $\epsilon^*$ to $\tilde{\epsilon}$ through $\mathcal{J}(\epsilon)$, we uncover the non-monotonic dependence of the equilibrium inspection intensity $m^*$ on $\eta$ (Proposition~\ref{The:intensity_deletion}). We present the key results below and defer the technical details to Appendix~J.

First, we present the AI operator's strategic MU decision in response to increasing $\eta$.

\begin{lemma}\label{Prop:epsilon_deletion}
The AI operator's unlearning certification level $\epsilon^*$ monotonically increases with the unlearning data proportion $\eta$ in the equilibrium.
\end{lemma}

Lemma~\ref{Prop:epsilon_deletion} reveals that as the data deletion requests increase (i.e., as $\eta$ rises), the AI operator raises the certification level~$\epsilon^*$, thereby weakening unlearning guarantees and further relaxing MU compliance. Intuitively, as $\eta$ increases, the model suffers greater utility loss from extensive data deletion. To mitigate this loss, the operator compensates by reducing the extent of unlearning, which is reflected in a higher certification level $\epsilon^*$.

\begin{figure}[t]
    \centering 
    \vspace{-15pt}
    \includegraphics[width=0.8\linewidth]{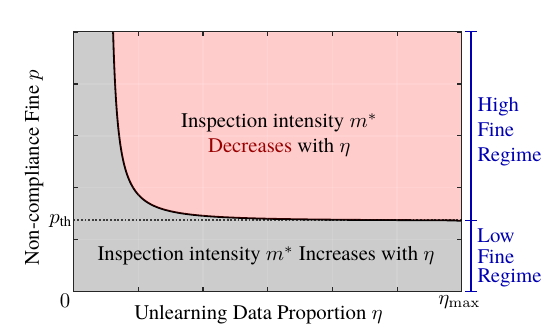}
    \vspace{-5pt}
    \caption{Illustration of how the auditor adjusts equilibrium inspection intensity $m^*$ in response to increasing unlearning data proportion $\eta$. Here, the boundary that separates the gray and red regions is defined by $\eta_{\text{th}}(p)$ in \eqref{equ:threshold_xi1}.}
    \label{fig:auditor_paradox}
    \vspace{-10pt}
\end{figure}

Next, we explore the auditor's equilibrium inspection intensity $m^*$. Specifically, we characterize how the auditor adjusts $m^*$ as $\eta$ rises across the $(\eta, p)$-plane, as formalized in Proposition~\ref{The:intensity_deletion} and Fig.~\ref{fig:auditor_paradox}. For clarity, we introduce two key thresholds: the~\textit{non-compliance fine threshold} $p_{\text{th}}$, and the \textit{unlearning data proportion threshold} $\eta_{\text{th}}(p)$:
\begin{equation}\label{equ:threshold_xi1}
p_{\text{th}}\triangleq\frac{ecG{\eta_{\max}}^2}{\left(2G{\eta_{\max}}^2-ec\right)\epsilon_0},\ \text{and}\ \eta_{\text{th}}(p)\triangleq \sqrt{\frac{pec\epsilon_0}{\left(2p\epsilon_0-ec\right)G}},
\end{equation}
where $\eta_{\text{th}}(p)$ decreases with the non-compliance fine $p$.\footnote{For the simplicity of presentation, we assume $2G{\eta_{\max}}^2>ec$ throughout this analysis, ensuring both regimes in Proposition \ref{The:intensity_deletion}. Relaxing this assumption yields only the high-fine regime, though our major insights will remain valid.}

\begin{proposition}\label{The:intensity_deletion}
The impact of unlearning data proportion $\eta$ on the auditor's equilibrium inspection intensity $m^*$ is illustrated in Fig. \ref{fig:auditor_paradox} and summarized as follows.
\begin{itemize}
    \item \mbox{\textbf{\emph{High-fine Regime with}} $p>p_\text{th}$\emph{\textbf{:}}} The inspection intensity $m^*$ is non-monotonic in $\eta$. Specifically, it increases with $\eta$ over $[0,\eta_{\text{th}}(p)]$ and then decreases over $(\eta_{\text{th}}(p),\eta_{\max}]$.
    \item \mbox{\textbf{\emph{Low-fine Regime with}} $p\le p_\text{th}$\emph{\textbf{:}}} The inspection intensity $m^*$ increases monotonically with $\eta$ over $[0,\eta_{\max}]$.
\end{itemize}
\end{proposition}

Proposition~\ref{The:intensity_deletion} highlights an unlearning audit paradox: simply intensifying inspections in response to rising deletion requests is not always optimal. This counterintuitive pattern arises when data deletion requests are large and the non-compliance fine~is high, as highlighted in the red region of Fig.~\ref{fig:auditor_paradox}. Particularly, as $\eta$ increases, the AI operator raises the certification level $\epsilon^*$ (see Lemma \ref{Prop:epsilon_deletion}), thereby relaxing unlearning stringency and making non-compliance easier to detect. Consequently, the auditor can reduce inspection intensity $m^*$ to cut costs while the detection probability $\text{Detect}(m^*,\epsilon^*)$ nonetheless improves.

On the other hand, when data deletion requests are relatively few, or the non-compliance fine is low, intensifying audits~$m^*$ as $\eta$ increases remains optimal, as illustrated in the gray region of Fig.~\ref{fig:auditor_paradox}. But the mechanisms behind differ across regimes:

\begin{itemize}
    \item \textit{High-fine regime with small $\eta$ (see the upper gray region):} 
    For a given fine $p$, when the unlearning proportion $\eta$ falls below the threshold $\eta_{\text{th}}(p)$, the AI operator can maintain a low certification level~$\epsilon^*$ and unlearn more effectively (see Lemma \ref{Prop:epsilon_deletion}). To uncover the resulting non-compliance, the auditor must raise inspection intensity~$m^*$ as~$\eta$ increases.

    \item \mbox{\textit{Low-fine regime (see the lower gray region):}} Since each detected non-compliance generates only limited fines, the auditor must consistently increase inspection intensity~$m^*$ as~$\eta$ grows to generate sufficient revenue to offset auditing costs.
\end{itemize}

Finally, we summarize the key finding from Proposition~\ref{The:intensity_deletion} below:

\begin{observation}[Unlearning Audit Paradox]\label{obs:audit_paradox}
The auditor may optimally reduce inspection intensity as data deletion requests increase when the deletion volume and the fine are high.
\end{observation}

Having established these theoretical results, we next shift~to real-world contexts to showcase the practical relevance of our findings.

\subsection{Real-world Illustrative Comparison: EU versus China}\label{subsec:comparative}

In this subsection, we compare the regulatory environments and auditing practices in the EU and China to contextualize our theoretical analysis. To be specific, we illustrate how regional differences in fine severity and deletion request volumes relate to observed auditing practices, showing qualitative consistency with our model's predictions from Proposition~\ref{The:intensity_deletion}~(see Fig.~\ref{fig:auditor_paradox}).

\subsubsection{EU: Lower Deletion Volumes, Milder Fines, and Intensified Audits (Gray Region of Fig.~\ref{fig:auditor_paradox})} After the RTBF~was~first established in 2014 and reinforced by the GDPR in~2018,~the volume of data deletion requests in the EU initially surged~but has diminished to a relatively low level since 2022 \cite{a2024_which}. Meanwhile, regulatory enforcement in the EU remains lenient,~with fines imposed infrequently and usually reduced afterward. For instance, several DPAs have issued reprimands instead of fines, even for confirmed non-compliance \cite{giles_2019_data}. From 2018 to 2023, only 1.3\% of the EU DPA investigations resulted in fines~\cite{a2025_data}, many of which were later significantly reduced \cite {a2020_ico, gabel_2021_germany}. 

As such, this regulatory environment, characterized by lower deletion volumes and milder fines, positions the~EU~in~the~gray region of Fig.~\ref{fig:auditor_paradox}, wherein our model suggests that the auditor should increase inspection intensity $m^*$ as $\eta$ increases. Qualitatively consistent with this theoretical prediction, several EU countries (e.g., the Netherlands and Sweden) have ramped up auditing~in response to rising data deletion requests \cite{amini_2025_swedish, vasylyk_2024_data}.

\subsubsection{China: Higher Deletion Volumes, Steeper Fines, and~Reduced Audits (Red Region in Fig. \ref{fig:auditor_paradox})} 

As a more recent entrant to global data governance, China has adopted a notably stricter regulatory stance. China's PIPL, effective since 2021, imposes significantly steeper fines than the EU's GDPR. Furthermore, as one of the world's largest digital economies, China is expected to face substantial deletion request volumes following the implementation of PIPL, which is similar to the EU's early surge in public engagement after the initial RTBF ruling \cite{surfshark_2023_how}.

Accordingly, this regulatory environment, featured by higher deletion volumes and steeper fines, positions China in the red region of Fig.~\ref{fig:auditor_paradox}, wherein our model suggests that the auditor should reduce inspection intensity $m^*$ as $\eta$ increases. Qualitatively consistent with this theoretical prediction, China's 2025 compliance measures explicitly mandate less frequent audits of personal information handling \cite{huld_2025_chinas}.

\subsubsection{Summary} 
To summarize, this comparison of the EU and China yields the following two key insights:
\begin{itemize}
    \item [(a)] \textit{Interpreting divergent auditing practices:} Our theoretical model suggests one plausible rationalization for divergent regional auditing strategies: as deletion requests grow, the EU intensifies auditing while China reduces it, potentially reflecting regional differences in fine severity and deletion volumes.

    \item [(b)] \textit{Linking theory to regulatory practice:} Our theoretical~predictions exhibit qualitative consistency with publicly observed regulatory developments across different regions, suggesting that our model provides practical relevance for interpreting real-world policy dynamics.
\end{itemize}

\section{Impact of Auditing Transparency}\label{sec:disclosure}
\begin{figure*}
\vspace{-20pt}
\begin{equation}\label{equ:SPE_operator}
\epsilon^d(m^d)=\frac{m^dG\eta^2\epsilon_0+\sqrt{\left(m^d\right)^2G^2\eta^4\epsilon_0^2+4G\eta^2m^d\left(G\eta^2\epsilon_0+p\epsilon_0^2\right)}}{2m^d\left(G\eta^2+p\epsilon_0\right)}.
\end{equation}
\begin{equation}\label{disclosed_Q}
\mathcal{Q}(m^d)=\gamma_1\left(p \cdot \left(1-\exp\left(-m^d\cdot\epsilon^d(m^d)\right)\right) - m^dc\right)+\gamma_2\left(-\epsilon^d(m^d)\right), \text{where $\epsilon^d(m^d)$ is given in \eqref{equ:SPE_operator}.}\vspace{-5pt}
\end{equation}
\hrulefill
\vspace{-10pt}
\end{figure*}
Thus far, we have focused on an undisclosed-auditing model (Section \ref{sec:system_model}) wherein the auditor chooses the inspection intensity without the operator's knowledge. However, policymakers increasingly advocate greater transparency in auditing to build public trust, a trend already observed in traditional sectors such as supply chains \cite{cho2019combating}. This raises a crucial question: should the auditor disclose her inspection policy to enhance transparency in MU compliance auditing?

To address this question, in this section, we further propose an alternative disclosed-auditing framework, wherein the auditor announces the inspection intensity before the AI operator's MU decision. Our analysis reveals a surprising finding: \textit{while undisclosed auditing provides an informational advantage in enforcing MU compliance, it paradoxically reduces regulatory cost-effectiveness compared to disclosed auditing.} Importantly, this finding challenges prevalent undisclosed and unannounced auditing practices, suggesting that regulators should reconsider embracing greater transparency in the MU auditing to improve regulatory cost-effectiveness and demonstrate commitment to safeguarding data privacy and RTBF in the AI era.

Specifically, we first introduce the disclosed-auditing framework in Section~\ref{subsec:disclosed}. We then analyze the new equilibrium~in Section \ref{subsec:newequilibrium}.  Finally, we explore the impact of auditing transparency by comparing the undisclosed-auditing and disclosed-auditing models in Section \ref{subsec:impact}.

\subsection{System Model: A Disclosed-auditing Framework}\label{subsec:disclosed}

To start, we first formulate the disclosed-auditing framework as a two-stage Stackelberg game, wherein the auditor publicly announces the inspection intensity $m$ before the operator's MU decision. Unlike the undisclosed-auditing model in Section \ref{sec:system_model}, the inspection policy is now known to the operator, though the operator's certification level remains private to himself:
\begin{itemize}
    \item \mbox{\textbf{Stage I:}} The auditor determines and publicly announces the inspection intensity $m$.
    \item \mbox{\textbf{Stage II:}} The AI operator decides the unlearning certification level $\epsilon$.
\end{itemize}
The subsequent auditing process follows the same procedure as the undisclosed-auditing model (see Fig.~\ref{fig:Deletion_Compliance_Model} and Section~\ref{subsubsec:interim_audit}). Given the sequential game formulation, we adopt the subgame perfect equilibrium (SPE) as our solution concept~\cite{fudenberg1991game}.

\subsection{New Equilibrium Analysis}\label{subsec:newequilibrium}

In this subsection, we proceed to analyze the equilibrium of the disclosed-auditing model through backward induction~\cite{fudenberg1991game}. Different from our undisclosed-auditing model in Section~\ref{sec:system_model}, the coupling between the AI operator's unlearning certification level and the auditor's inspections no longer persists here. This is because, by disclosing the inspection intensity $m$, the auditor in Stage I can fully anticipate the operator's response in Stage II and strategically influence the operator's MU decision. 


Notice that the AI operator's Stage II optimization problem is the same as that in the original undisclosed-auditing model, analyzed in Section~\ref{subsubsec:AI}. Therefore, we only need to analyze the auditor's optimal inspection intensity in Stage~I. Hereafter, we adopt the superscript $d$ to denote the equilibrium outcome of the disclosed-auditing model. In the following proposition, we provide the complete equilibrium characterization for this model. 

\begin{proposition}\label{Prop:SPE}
In the disclosed-auditing model, there exists a unique SPE as follows.
\begin{itemize}
    \item In Stage I, the auditor sets the optimal inspection intensity $m^d$ as the unique solution to $ \partial \mathcal{Q}(m^d)/\partial m^d = 0$,
    where $\mathcal{Q}(m^d)$ is defined in \eqref{disclosed_Q}.
    \item In Stage II, the AI operator's equilibrium unlearning~certification level $\epsilon^d(m^d)$ is given in \eqref{equ:SPE_operator}.
\end{itemize}
\end{proposition}

In contrast to the undisclosed-auditing model (see Theorem \ref{Theorem:NE}), the disclosed-auditing model features a distinct equilibrium characterization for the auditor's optimal inspection intensity. Specifically, by announcing $m^d$ before the operator's MU decision, the auditor's optimization now incorporates the operator's strategic Stage II response, as characterized in \eqref{disclosed_Q}.

\subsection{Impact of Auditing Transparency}\label{subsec:impact}

To better understand the impact of auditing transparency, we now compare the equilibrium behaviors and welfare outcomes across the undisclosed-auditing and disclosed-auditing models. 

First, we investigate the equilibrium inspection intensity and the corresponding certification level in the following corollary.

\begin{corollary}\label{coro:SPE}
Compared with the undisclosed-auditing model in Section~\ref{sec:system_model}, the AI operator's equilibrium unlearning certification level is higher and the auditor's equilibrium inspection intensity is lower in the disclosed-auditing model, i.e., 
\begin{equation}
    \epsilon^d> \epsilon^*\quad\text{ and }\quad m^d< m^*,
\end{equation}
if and only if the weight assigned to regulatory surplus is sufficiently large relative to social concern over MU compliance, i.e.,  $\gamma_1/\gamma_2>\gamma_{\mathrm{th}}$.\footnote{Here, the threshold $\gamma_{\mathrm{th}}(\epsilon^*,\epsilon^d)$ is characterized in Appendix~L.\label{footnote}}
\end{corollary}

Corollary~\ref{coro:SPE} indicates that greater auditing transparency can lead to weaker MU compliance (i.e., higher $\epsilon^d$). The intuition is that the disclosure permits the auditor to commit ex ante to a lower inspection intensity $m^d$, which relaxes the AI operator's incentive to adopt a strong unlearning certificate.

Next, we compare the auditor's expected payoff in the equilibrium across the two models in the following proposition.

\begin{proposition}\label{Prop:payoff}
(i) Compared with the undisclosed-auditing model in Section~\ref{sec:system_model}, the auditor attains a strictly higher expected regulatory surplus under the disclosed-auditing model, i.e.,
\begin{equation}
    \Pi^d_c>\Pi^*_c
\end{equation}
if and only if the weight assigned to regulatory surplus is sufficiently large relative to social concern over MU compliance, i.e.,  $\gamma_1/\gamma_2>\gamma_{\mathrm{th}}$.\footref{footnote}

(ii) Moreover, regardless of the relative weights $\gamma_1$ and $\gamma_2$, the auditor's total expected payoff is always higher under the disclosed-auditing model, i.e.,
\begin{equation}
\quad\Pi^d\ge\Pi^*.
\end{equation}
\end{proposition}

Proposition~\ref{Prop:payoff} reveals a surprising result: disclosed auditing improves regulatory cost-effectiveness despite undisclosed auditing's informational advantage in enforcing MU compliance. The underlying mechanism is the auditor's commitment power through the disclosure of the auditing information. By publicly announcing inspection intensity in Stage I, the auditor can credibly commit to a lower inspection intensity $m^d$. This commitment mitigates
over-auditing: a lower $m^d$ yields substantial regulatory surplus through both auditing cost savings and more fine revenue, while accepting slightly weaker MU compliance (Corollary~\ref{coro:SPE}). We summarize this finding below:

\begin{observation}[\mbox{Auditing Transparency Paradox}]\label{obs:transparency_paradox}
Although undisclosed auditing provides an informational advantage in enforcing MU compliance, it paradoxically reduces regulatory cost-effectiveness compared to disclosed auditing.
\end{observation}

Therefore, policymakers should reconsider the widespread use of undisclosed auditing in existing AI and data governance frameworks. In contrast, disclosed auditing not only enhances cost-effectiveness but also builds public trust by demonstrating regulators' commitment to safeguarding privacy in the AI era.


\section{Simulations}\label{sec:experiment}

\begin{figure}
    \vspace{-5pt}
    \centering
    \includegraphics[width=0.7\linewidth]{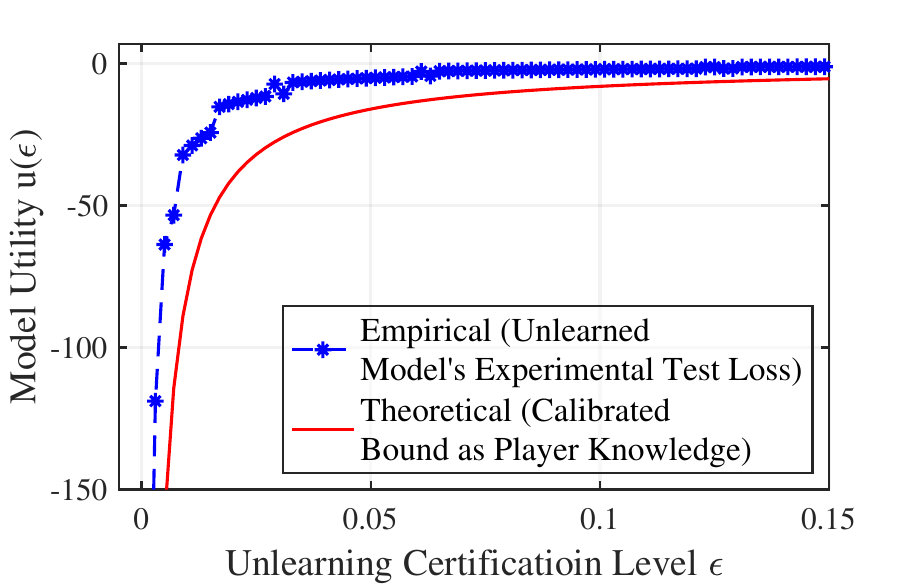}
    \vspace{-5pt}
    \caption{Theoretical-empirical Alignment: The model utility $u(\epsilon)$, derived from the empirical test loss and the calibrated theoretical upper bound based~on~\eqref{equ:model_loss_simplify}, versus the unlearning certification level $\epsilon$ (illustrated for $\eta=0.02$). Here,~the model characteristic parameter is calibrated to $G=2,000$ to guarantee~that the calibrated theoretical curve provides a consistent bound on the empirical test loss across practical parameter settings.}
    \label{fig_theoretical_empirical}
    \vspace{-10pt}
\end{figure}
In this section, we conduct real-data-based experiments and simulations to validate our analytical findings and evaluate the performance of our proposed mechanisms. In Section~\ref{subsec:experiment_setup}, we introduce the experimental setting and validate theoretical-empirical alignment through experiments, which help calibrate the model characteristic parameter $G$ that will serve as critical auditing information. Based on this calibration, Section~\ref{subsec:experiment_NE} simulates the equilibrium behaviors of both players, revealing several regulatory implications. Finally, we evaluate the performance of our proposed mechanisms against a state-of-the-art benchmark in Section \ref{subsec:perf}.

\subsection{Experimental Setup and Empirical Validation}\label{subsec:experiment_setup}

\begin{figure*}[t]
\vspace{-25pt}
    \subfigure[$\epsilon^*$ versus $p$]{
        \begin{minipage}{0.31\linewidth}
            \centering
            \includegraphics[width=1\linewidth]{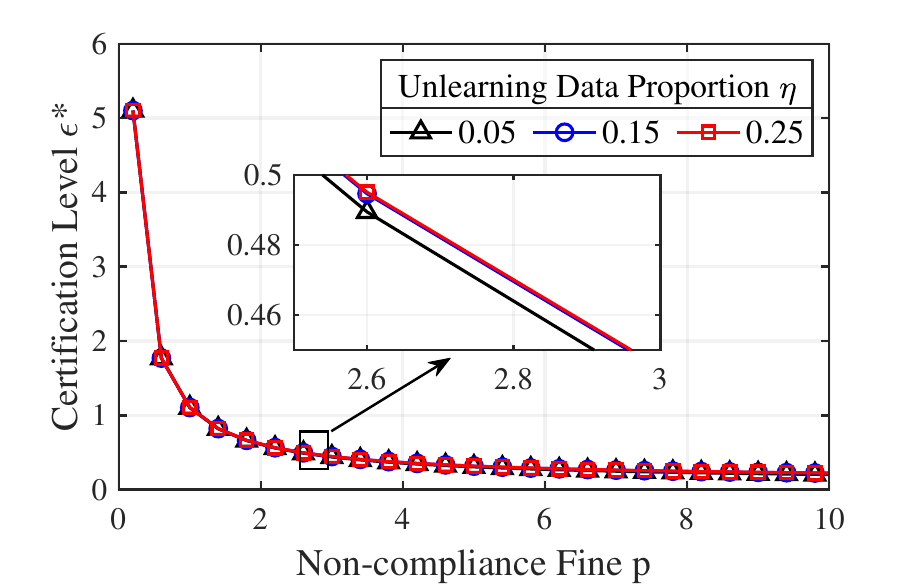}
            \vspace{-5pt}
            \label{fig_penalty}
		\end{minipage}
	}
    \subfigure[$\epsilon^*$ versus $\eta$]{
        \begin{minipage}{0.31\linewidth}
            \centering
            \includegraphics[width=1\linewidth]{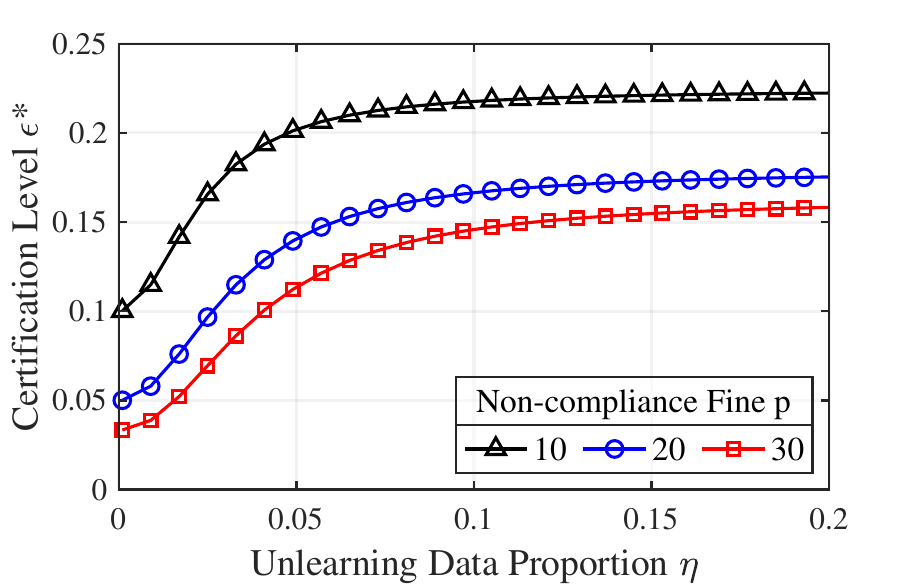}
            \vspace{-5pt}
            \label{fig_cert_eta}
		\end{minipage}
	}
	\subfigure[$m^*$ versus $\eta$]{
        \begin{minipage}{0.31\linewidth}
            \centering
            \includegraphics[width=1\linewidth]{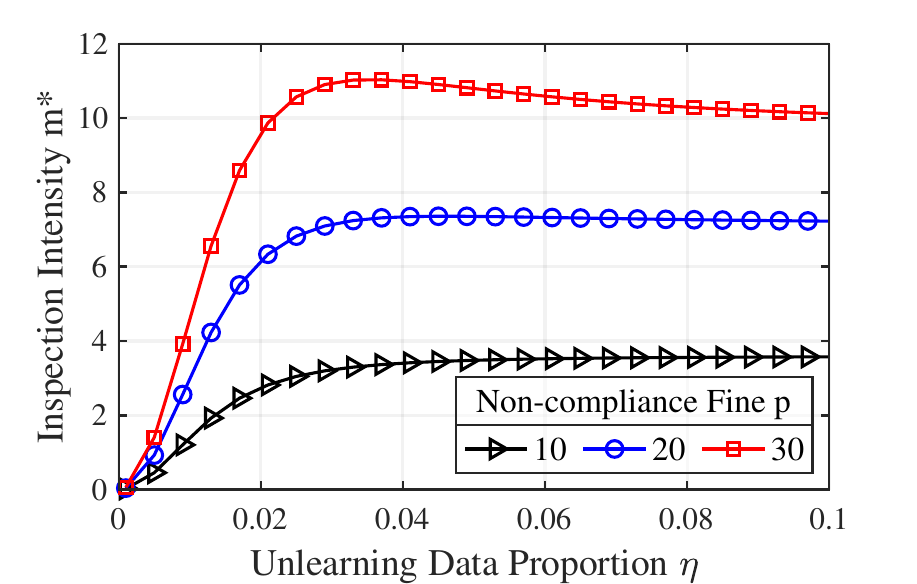}
            \vspace{-5pt}
            \label{fig_audit_eta}
		\end{minipage}
	}\vspace{-5pt}
    \caption{(a) The AI operator's unlearning certification level $\epsilon^*$ versus the non-compliance fine $p$ in the NE. (b) The AI operator's unlearning certification level $\epsilon^*$ and (c) the auditor's inspection intensity $m^*$ versus the unlearning data proportion $\eta$ in the NE. Here, in Fig. \ref{fig_penalty}, we vary the unlearning data proportion $\eta$ over $\{0.05,0.15,0.25\}$ for each curve, and in Figs. \ref{fig_cert_eta} and \ref{fig_audit_eta}, we vary the non-compliance fine $p$ over $\{10,20,30\}$ for each curve.}
    \vspace{-10pt}
\end{figure*}

\subsubsection{Experiment Setup}\label{subsubsec:setup}

Following \cite{qiao2024hessian}, we use a multinomial logistic regression model with $7,850$ parameters for handwritten digit classification, trained on $1,000$ samples and evaluated on $400$ test samples all sourced from the MNIST dataset~\cite{deng2012mnist}. Training employs the cross-entropy loss with $L^2$-regularization (coefficient $0.5$) over $15$ epochs, with a learning rate of $0.05$, decay rate of $0.995$, and batch size of $64$. This model will serve as the baseline for performing MU operations to accommodate data deletion requests. For a given unlearning data proportion $\eta$, we generate the unlearned model through the Hessian-free MU method from~\cite{qiao2024hessian}.

Regarding the system parameter of the auditing framework, we set the auditing confidence level to $\xi = 0.99$, the minimum deletion level for corrective unlearning to $\epsilon_0 = 0.1$, and the unit inspection cost to $c = 1$. In the auditor's payoff in \eqref{Equ:payoff_auditor}, the weights assigned to the regulatory surplus and social concern over MU compliance are $\gamma_1 = 0.5$ and $\gamma_2 = 0.5$, respectively.

\subsubsection{Framework Validation and Empirical Calibration}\label{subsubsec:frame_validation}

Next, we conduct real-data MU experiments to obtain the empirical test loss of the unlearned model, which would help validate~the theoretical-empirical alignment of our framework and calibrate the model characteristic parameter $G$. 

Specifically, we calibrate the model characteristic parameter to $G = 2,000$ to ensure the calibrated theoretical bound based on \eqref{equ:model_loss_simplify} provides a consistent upper bound on the empirical test loss across practical parameter settings. Fig. \ref{fig_theoretical_empirical} illustrates this calibration for $\eta = 0.02 $. The model utility $ u(\epsilon) $ derived from the unlearned model's empirical test loss is shown by the blue dotted curve, whereas the red solid curve represents the model utility derived from the theoretical bound using the calibrated $G = 2,000 $. Due to space limitations, we here present only this specific calibration instance, with additional validation results from our extensive experiments provided in Appendix N.\footnote{As discussed in Section~\ref{subsubsec:model_utility_function}, $G$ consolidates model-specific quantities that vary across MU methods and are difficult to directly obtain in practice. We calibrate $G=2,000$ to demonstrate qualitative alignment between theoretical and empirical trends. Our focus here is on establishing behavioral consistency and deriving policy insights rather than precise quantitative calibration, which would require complex model-specific experiments.}

Moreover, Fig.~\ref{fig_theoretical_empirical} demonstrates strong qualitative alignment: both curves exhibit the predicted inverse relationship between $u(\epsilon)$ and $\epsilon$, with a concavely increasing pattern, as described in \eqref{equ:model_loss_simplify}. This also confirms that our framework effectively captures the empirical behavior of unlearned models. Importantly, this calibrated theoretical model utility will serve as essential audit information, enabling both players to make informed strategic decisions based on an understanding of model characteristics.

\subsection{Equilibrium Outcomes and Their Implications}\label{subsec:experiment_NE}

Next, based on the calibrated model characteristic parameter $G$ from Fig.~\ref{fig_theoretical_empirical}, we simulate the equilibrium behaviors of both the AI operator and the auditor to investigate their implications for regulatory auditing. In this subsection, we concentrate on the undisclosed-auditing model in Section~\ref{sec:system_model},  and will explore the disclosed-auditing model in the following subsection.

\subsubsection{Impact of Non-compliance Fine}\label{subsubsec:fine}

We begin by examining how the non-compliance fine regulates MU compliance in the equilibrium. As shown in Fig. \ref{fig_penalty}, the operator's equilibrium unlearning certification level $\epsilon^*$ decreases convexly in the fine $p$ across various unlearning data proportions $\eta$: it drops sharply for small $p$ and gradually levels off beyond a threshold (e.g., $p\gtrsim 2$). This highlights the penalty saturation effect, consistent with Corollary \ref{Cor:BR_certification_penalty} from the AI operator's best-response analysis, confirming its persistence in equilibrium. This effect carries an important policy implication: excessively high penalties can be counterproductive, imposing social costs without proportional compliance gains. This insight aligns with real-world privacy regulations, such as the GDPR \cite{gdpr_2018_general} and PIPL \cite{thena}, which cap the RTBF non-compliance fine. We formalize this finding below:

\begin{observation}[Penalty Saturation Effect]\label{obs:penalty_saturation_effect}
Beyond a certain threshold, further increases in the non-compliance fine yield diminishing returns in MU compliance improvement.
\end{observation}

\subsubsection{Impact of Unlearning Data Proportion}\label{subsubsec:volume}

Next, we study how the unlearning data proportion~$\eta$ impacts the equilibrium behaviors of both the AI operator and the auditor, respectively.

As shown in Fig. \ref{fig_cert_eta}, the AI operator consistently increases the certification level $\epsilon^*$ as $\eta$ increases to mitigate the growing model utility loss from MU. This is consistent with Lemma~\ref{Prop:epsilon_deletion}. However, an MU compliance plateau effect emerges at high~$\eta$: $\epsilon^*$ initially increases but gradually stabilizes, reflecting diminishing returns from relaxed compliance as $\eta$ grows. This seems counterintuitive: a higher $\eta$ results in greater model utility loss, suggesting the operator should increase $\epsilon^*$ more aggressively. However, once $\epsilon^*$ is high, further increases yield diminishing marginal utility gains (Fig.~\ref{fig_theoretical_empirical}) that cannot offset the rising audit risk. We formalize this finding below:

\begin{observation}[Compliance Plateau Effect]
As data deletion requests increase, relaxing MU compliance beyond a certain point yields diminishing returns for the AI operator.
\end{observation}

As shown in Fig. \ref{fig_audit_eta}, the auditor's inspection intensity~$m^*$ decreases as $\eta$ increases when both the non-compliance fine $p$ and the unlearning data proportion $\eta$ are high (i.e., $\eta>0.0370$ for $p=30$ and $\eta>0.0450$ for $p=20$). This finding validates the unlearning audit paradox in Proposition \ref{The:intensity_deletion} and Observation \ref{obs:audit_paradox}. In addition, across various fines $p$, the auditor's adjustments to the inspection intensity $m^*$ become minor at high $\eta$. This naturally connects to the compliance plateau effect in Fig. \ref{fig_cert_eta}: at high $\eta$, the AI operator stabilizes MU compliance, requiring merely minimal inspection adjustments to maintain the optimal regulatory cost-effectiveness. We formalize this finding below:

\begin{observation}[Auditing Saturation Effect]\label{obs:audit_saturation}
Beyond a certain threshold, adjusting inspection intensity yields diminishing returns for the auditor as data deletion requests increase.
\end{observation}

\subsection{Performance Evaluation}\label{subsec:perf}

\begin{figure}[t]
    \centering
    \vspace{-13pt}
    \subfigure[The Auditor's Expected Payoff]{
        \begin{minipage}{0.465\linewidth}
            \centering
            \includegraphics[width=1.05\linewidth]{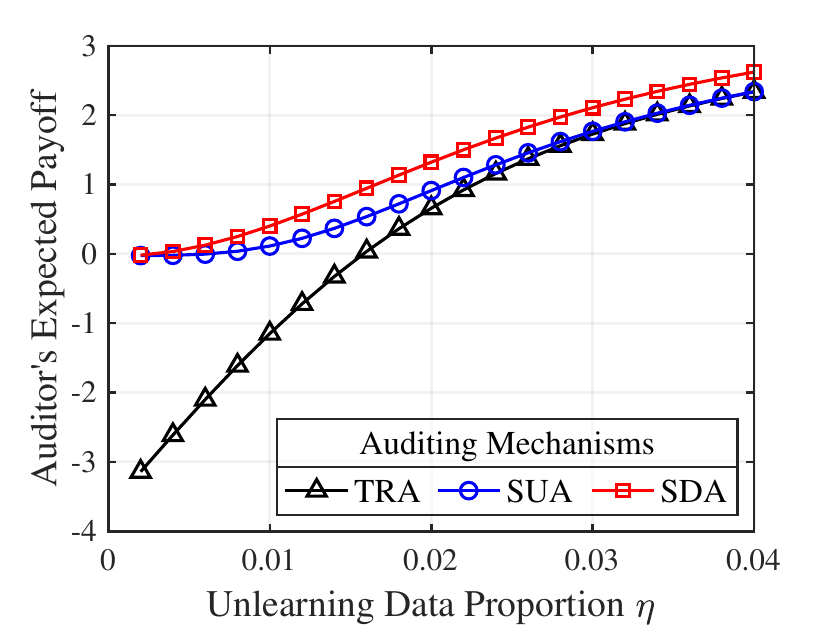}
            \vspace{-2.5pt}
            \label{fig_perf_audit}
		\end{minipage}
	}
    \subfigure[AI Operator's Empirical Payoff]{
        \begin{minipage}{0.465\linewidth}
            \centering
            \includegraphics[width=1.05\linewidth]{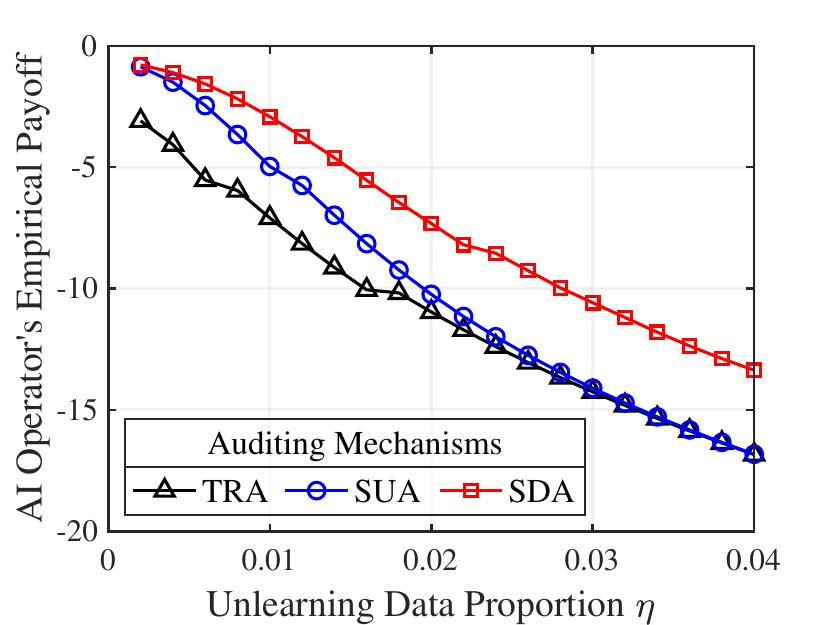}
            \vspace{-2.5pt}
            \label{fig_perf_AI}
		\end{minipage}
	}
\caption{Performance comparisons in terms of (a) the auditor's expected payoff (which measures regulatory cost-effectiveness) in the equilibrium and (b) the AI operator's empirical payoff in the equilibrium, under the three mechanisms (TRA, SUA, SDA). Here, the non-compliance fine is set to $p=20$.}
\label{fig_perf}
\vspace{-10pt}
\end{figure}

Finally, we evaluate the performance of our proposed mechanisms against a benchmark adapted from existing literature: 
\begin{itemize}
    \item \textit{Traditional Risk-based Auditing (TRA)} \cite{sommer2020towards,wang2024has}: The auditor designs the inspection intensity merely based on the estimated non-compliance risk and auditing costs, without considering the AI operator's strategic MU decision.\footnote{Owing to space limitations, we defer the detailed analysis and closed-form solutions for the optimal inspection intensity and unlearning certification level under the TRA benchmark to Appendix~O.}
    \item \textit{Our Proposed Strategic Undisclosed Auditing (SUA)}: The auditor designs the inspection intensity while accounting for the AI operator's strategic MU decision, but keeps the inspection intensity undisclosed.
    \item \textit{Our Proposed Strategic Disclosed Auditing (SDA)}: The auditor designs the inspection intensity while accounting for the AI operator's strategic MU decisions and publicly discloses the inspection intensity in advance.
\end{itemize}
Fig.~\ref{fig_perf_audit} and Fig.~\ref{fig_perf_AI} compare the performance of these three auditing mechanisms in terms of the auditor's expected payoff (which measures the regulatory cost-effectiveness) and the AI operator's empirical payoff in the equilibrium, respectively.\footnote{In Fig. \ref{fig_perf_AI}, we conduct real-data MU experiments to obtain the empirical test loss of the unlearned model and then compute the corresponding payoff.}

As demonstrated in Figs.~\ref{fig_perf_audit} and~\ref{fig_perf_AI}, our~proposed~SUA and SDA mechanisms outperform the TRA benchmark in both the auditor's and the AI operator's payoffs. Compared to TRA, the maximum improvements achieved by SUA and SDA reach $1404.84\%$ and $2549.30\%$ for the auditor, whereas $72.09\%$ and $74.60\%$ for the AI operator, respectively. By accounting for the AI operator's strategic MU decision, our proposed mechanisms allow the auditor to choose the inspection intensity more cost-effectively, with these improvements most pronounced at small unlearning data proportions (e.g., $\eta\le 0.02$). However, as $\eta$ increases, the operator's certification level gradually stabilizes (i.e., the compliance plateau effect in Fig. \ref{fig_cert_eta}), and therefore the benefits of gauging strategic MU non-compliance diminish, which echoes the auditing saturation effect in Fig. \ref{fig_audit_eta}.

More importantly, the SDA mechanism consistently outperforms the SUA mechanism for both parties. For the auditor, SDA improves regulatory cost-effectiveness by up to $562.69\%$, which empirically validates Proposition \ref{Prop:payoff}. For the AI operator, SDA improves his payoff by up to $41.25\%$. Accordingly, these findings demonstrate that auditing transparency creates a win-win outcome for both parties. We formalize this insight below:

\begin{observation}[\mbox{Win-win Solution of Auditing Transparency}]
Disclosed auditing outperforms undisclosed auditing in terms of both players' payoffs, creating a win-win solution through enhanced auditing transparency.
\end{observation}

\section{Conclusion}\label{sec:conclusion}
This paper presents the first economic framework for auditing MU compliance, addressing a significant gap in emerging AI and data-governance practices. By integrating the certified unlearning theory with regulatory enforcement, we model and analyze the strategic interactions between the AI operator and the auditor. Despite the technical challenges posed by the~MU-induced nonlinearities, we employ auxiliary transformations to decouple the system and to establish the existence, uniqueness, and key structural properties of the equilibrium without relying on explicit closed-form solutions.~Our analysis highlights that, as deletion requests increase, the auditor can even optimally reduce inspection intensity because less strict unlearning makes non-compliance easier to detect. Moreover, we also prove that although undisclosed auditing offers informational advantages, it paradoxically reduces regulatory cost-effectiveness relative to disclosed auditing. Furthermore, disclosed auditing creates a win-win outcome for both players through enhanced auditing transparency. Experimental results demonstrate that, compared to the state-of-the-art benchmark, disclosed auditing increases the auditor's payoff by up to $2549.30\%$ and the AI operator's payoff by up to $74.60\%$, with this advantage most pronounced at small unlearning data proportions.

Future research could explore how auditors adaptively learn regulatory parameters (e.g., model characteristics and deletion volume distributions) through repeated interactions with the AI operator to optimize the inspection strategies. Another~promising direction is to develop richer auditing frameworks that accommodate heterogeneity among AI operators and incorporate more flexible detection rules and penalty schemes. Finally, we plan to pursue empirical validations by implementing practical MU verification procedures in large-scale AI systems, further strengthening the applicability and insights of our framework.

\bibliographystyle{IEEEtran}
\bibliography{sample-base}

@String{Computing = "Computing" }

@String{Computer = "{IEEE} Computer" }

@inproceedings{bertram2019five,
  title={Five years of the right to be forgotten},
  author={Bertram, Theo and Bursztein, Elie and Caro, Stephanie and Chao, Hubert and Chin Feman, Rutledge and Fleischer, Peter and Gustafsson, Albin and Hemerly, Jess and Hibbert, Chris and Invernizzi, Luca and others},
  booktitle={Proceedings of the 2019 ACM SIGSAC Conference on Computer and Communications Security},
  pages={959--972},
  year={2019}
}

@article{sekhari2021remember,
  title={Remember what you want to forget: Algorithms for machine unlearning},
  author={Sekhari, Ayush and Acharya, Jayadev and Kamath, Gautam and Suresh, Ananda Theertha},
  journal={Advances in Neural Information Processing Systems},
  volume={34},
  pages={18075--18086},
  year={2021}
}

@article{wang2024has,
  title={Has Approximate Machine Unlearning been evaluated properly? From Auditing to Side Effects},
  author={Wang, Cheng-Long and Li, Qi and Xiang, Zihang and Wang, Di},
  journal={arXiv preprint arXiv:2403.12830},
  year={2024}
}

@article{sommer2020towards,
  title={Towards probabilistic verification of machine unlearning},
  author={Sommer, David Marco and Song, Liwei and Wagh, Sameer and Mittal, Prateek},
  journal={arXiv preprint arXiv:2003.04247},
  year={2020}
}

@book{fudenberg1991game,
  title={Game theory},
  author={Fudenberg, Drew and Tirole, Jean},
  year={1991},
  publisher={MIT press}
}

@article{finley1994game,
  title={Game theoretic analysis of discovery sampling for internal fraud control auditing},
  author={Finley, David R},
  journal={Contemporary Accounting Research},
  volume={11},
  number={1},
  pages={91--114},
  year={1994},
  publisher={Wiley Online Library}
}

@article{cho2019combating,
  title={Combating child labor: Incentives and information disclosure in global supply chains},
  author={Cho, Soo-Haeng and Fang, Xin and Tayur, Sridhar and Xu, Ying},
  journal={Manufacturing \& Service Operations Management},
  volume={21},
  number={3},
  pages={692--711},
  year={2019},
  publisher={INFORMS}
}

@inproceedings{kairouz2015composition,
  title={The composition theorem for differential privacy},
  author={Kairouz, Peter and Oh, Sewoong and Viswanath, Pramod},
  booktitle={International conference on machine learning},
  pages={1376--1385},
  year={2015},
  organization={PMLR}
}

@misc{gdpr_art,
  title = {Article 58: Powers},
  url = {https://gdpr-info.eu/art-58-gdpr/},
  urldate = {2025-04},
  year = {2018},
  organization = {General Data Protection Regulation (GDPR)}
}

@inproceedings{zhang2024verification,
  title={Verification of machine unlearning is fragile},
  author={Zhang, Binchi and Chen, Zihan and Shen, Cong and Li, Jundong},
  booktitle={Proceedings of the 41st International Conference on Machine Learning},
  pages={58717--58738},
  year={2024}
}

@inproceedings{thudi2022necessity,
  title={On the necessity of auditable algorithmic definitions for machine unlearning},
  author={Thudi, Anvith and Jia, Hengrui and Shumailov, Ilia and Papernot, Nicolas},
  booktitle={31st USENIX security symposium (USENIX Security 22)},
  pages={4007--4022},
  year={2022}
}

@misc{surfshark_2023_how,
  title = {How many people in Europe use their “right to be forgotten” online?},
  url = {https://surfshark.com/blog/right-to-be-forgotten-requests},
  urldate = {2025-05},
  year = {2023},
  organization = {Surfshark}
}

@misc{a2024_which,
  title = {Which countries exercise the “right to be forgotten” the most?},
  url = {https://surfshark.com/research/study/right-to-be-forgotten},
  urldate = {2025-05},
  year = {2024},
  organization = {Surfshark}
}

@misc{huld_2025_chinas,
  author = {Huld, Arendse},
  title = {China's Personal Information Protection Audit: A Guide for Businesses},
  url = {https://www.china-briefing.com/news/chinas-personal-information-protection-audit-final-measures-effective-may-1/},
  urldate = {2025-07},
  year = {2025},
  organization = {China Briefing News}
}

@misc{giles_2019_data,
  author = {Giles, John},
  title = {Data protection audit by an authority: GDPR audit},
  url = {https://www.michalsons.com/blog/data-protection-audit-by-an-authority-gdpr-audit-2/40732},
  urldate = {2025-05},
  year = {2019},
  organization = {Michalsons}
}

@article{yu2017public,
  title={Public Wi-Fi monetization via advertising},
  author={Yu, Haoran and Cheung, Man Hon and Gao, Lin and Huang, Jianwei},
  journal={IEEE/ACM Transactions on Networking},
  volume={25},
  number={4},
  pages={2110--2121},
  year={2017},
  publisher={IEEE}
}

@article{fainmesser2023digital,
  title={Digital privacy},
  author={Fainmesser, Itay P and Galeotti, Andrea and Momot, Ruslan},
  journal={Management Science},
  volume={69},
  number={6},
  pages={3157--3173},
  year={2023},
  publisher={INFORMS}
}

@misc{gobert_2024_audit,
  author = {Gobert, Xavier and Camberlin, Emeraude},
  title = {Audit by a Data Protection Authority: How to Be Prepared?},
  url = {https://onenucleus.com/audit-data-protection-authority-how-be-prepared},
  urldate = {2025-03},
  year = {2024},
  organization = {Onenucleus}
}

@misc{a2022_practical,
  organization = {The French Data Protection Authority (CNIL).},
  title = {Practical Guide to GDPR for Data Protection Officers},
  url = {https://www.cnil.fr/sites/cnil/files/atoms/files/cnil-gdpr_practical_guide_data-protection-officers.pdf},
  urldate = {2025-03},
  year = {2022}
}

@misc{vasylyk_2024_data,
  author = {Vasylyk, Olya},
  title = {Data protection digest: Sneakily changing terms of service and privacy policy won't help your business},
  url = {https://techgdpr.com/blog/data-protection-digest-19022024-sneakily-changing-terms-of-service-and-privacy-policy-wont-help-your-business/},
  urldate = {2025-03},
  year = {2024},
  organization = {TechGDPR}
}

@article{qiao2024hessian,
  title={Hessian-Free Online Certified Unlearning},
  author={Qiao, Xinbao and Zhang, Meng and Tang, Ming and Wei, Ermin},
  journal={arXiv preprint arXiv:2404.01712},
  year={2024}
}

@misc{schwartz_2021_gdpr,
  author = {Schwartz, Samantha},
  title = {GDPR regulators are sinking their teeth into violators. 2020's fines are proof.},
  url = {https://www.cybersecuritydive.com/news/data-privacy-regulator-gdpr-fines/594084/},
  urldate = {2025-04},
  year = {2021},
  organization = {Cybersecurity Dive}
}

@misc{a2020_guidance,
  organization = {Information Commissioner's Office (ICO)},
  title = {Guidance on the AI Auditing Framework},
  url = {https://ico.org.uk/media2/migrated/2617219/guidance-on-the-ai-auditing-framework-draft-for-consultation.pdf},
  urldate = {2025-11-21},
  year = {2020}
}

@article{suriyakumar2022algorithms,
  title={Algorithms that approximate data removal: New results and limitations},
  author={Suriyakumar, Vinith and Wilson, Ashia C},
  journal={Advances in Neural Information Processing Systems},
  volume={35},
  pages={18892--18903},
  year={2022}
}

@article{liu2023certified,
  title={Certified minimax unlearning with generalization rates and deletion capacity},
  author={Liu, Jiaqi and Lou, Jian and Qin, Zhan and Ren, Kui},
  journal={Advances in Neural Information Processing Systems},
  volume={36},
  pages={62821--62852},
  year={2023}
}

@misc{a2016_article,
  title = {Article 49: Registration},
  url = {https://artificialintelligenceact.eu/article/49/},
  urldate = {2025-11-20},
  year = {2024},
  organization = {EU Artificial Intelligence Act}
}

@article{deng2012mnist,
  title={The mnist database of handwritten digit images for machine learning research},
  author={Deng, Li},
  journal={IEEE signal processing magazine},
  volume={29},
  number={6},
  pages={141--142},
  year={2012},
  publisher={IEEE}
}

@inproceedings{guo2019certified,
  title={Certified data removal from machine learning models},
  author={Guo, Chuan and Goldstein, Tom and Hannun, Awni and Van Der Maaten, Laurens},
  booktitle={Proceedings of the 37th International Conference on Machine Learning},
  pages={3832--3842},
  year={2020}
}

@misc{a2020_ico,
  title = {ICO announces Significantly reduced GDPR fine for British Airways},
  url = {https://www.cliffordchance.com/insights/resources/blogs/talking-tech/en/articles/2020/10/ico-announces-significantly-reduced-gdpr-fine-for-british-airway.html},
  urldate = {2025-05-25},
  year = {2020},
  organization = {Clifford Chance}
}

@misc{gabel_2021_germany,
  author = {Gabel, Detlev and Langen, Markus},
  title = {Germany: Significant reduction of GDPR fines},
  url = {https://www.whitecase.com/insight-alert/germany-significant-reduction-gdpr-fines},
  urldate = {2025-05-25},
  year = {2021},
  organization = {White \& Case}
}

@misc{a2025_data,
  title = {Data Protection Day: Only 1.3\% of cases before EU DPAs result in a fine},
  url = {https://noyb.eu/en/data-protection-day-only-13-cases-eu-dpas-result-fine},
  urldate = {2025-05-25},
  year = {2025},
  organization = {NOYB - European Center for Digital Rights}
}

@article{becker1968crime,
  title={Crime and punishment: An economic approach},
  author={Becker, Gary S},
  journal={Journal of political economy},
  volume={76},
  number={2},
  pages={169--217},
  year={1968},
  publisher={The University of Chicago Press}
}

@article{tsebelis1990penalty,
  title={Penalty has no impact on crime: A game-theoretic analysis},
  author={Tsebelis, George},
  journal={Rationality and Society},
  volume={2},
  number={3},
  pages={255--286},
  year={1990},
  publisher={Sage Publications}
}

@article{plambeck2016supplier,
  title={Supplier evasion of a buyer’s audit: Implications for motivating supplier social and environmental responsibility},
  author={Plambeck, Erica L and Taylor, Terry A},
  journal={Manufacturing \& Service Operations Management},
  volume={18},
  number={2},
  pages={184--197},
  year={2016},
  publisher={INFORMS}
}

@inproceedings{bourtoule2021machine,
  title={Machine unlearning},
  author={Bourtoule, Lucas and Chandrasekaran, Varun and Choquette-Choo, Christopher A and Jia, Hengrui and Travers, Adelin and Zhang, Baiwu and Lie, David and Papernot, Nicolas},
  booktitle={2021 IEEE symposium on security and privacy},
  pages={141--159},
  year={2021},
}

@article{weng2024proof,
  title={Proof of unlearning: Definitions and instantiation},
  author={Weng, Jiasi and Yao, Shenglong and Du, Yuefeng and Huang, Junjie and Weng, Jian and Wang, Cong},
  journal={IEEE Transactions on Information Forensics and Security},
  volume={19},
  pages={3309--3323},
  year={2024},
  publisher={IEEE}
}

@ARTICLE{guo2023verifying,
  author={Guo, Yu and Zhao, Yu and Hou, Saihui and Wang, Cong and Jia, Xiaohua},
  journal={IEEE Transactions on Information Forensics and Security}, 
  title={Verifying in the Dark: Verifiable Machine Unlearning by Using Invisible Backdoor Triggers}, 
  year={2024},
  volume={19},
  pages={708-721},
}

@misc{fazlioglu_2023_iapp,
  author = {Fazlioglu, Müge},
  title = {Privacy and Consumer Trust Report},
  url = {https://iapp.org/resources/article/privacy-and-consumer-trust-summary/},
  urldate = {2025-11-20},
  year = {2023},
  organization = {The International Association of Privacy Professionals (IAPP)}
}

@misc{thena,
  title = {Personal Information Protection Law of the People's Republic of China},
  url = {http://en.npc.gov.cn.cdurl.cn/2021-12/29/c_694559.htm},
  urldate = {2025-11-20},
  year = {2021}
}

@inproceedings{cao2015towards,
  title={Towards making systems forget with machine unlearning},
  author={Cao, Yinzhi and Yang, Junfeng},
  booktitle={2015 IEEE symposium on security and privacy},
  pages={463--480},
  year={2015},
}

@misc{a2021_california,
  title = {California Company Settles FTC Allegations It Deceived Consumers about use of Facial Recognition in Photo Storage App},
  url = {https://www.ftc.gov/news-events/news/press-releases/2021/01/california-company-settles-ftc-allegations-it-deceived-consumers-about-use-facial-recognition-photo},
  urldate = {2025-07},
  year = {2021},
  organization = {Federal Trade Commission (FTC)}
}

@misc{_2024_statement,
  organization = {European Data Protection Board (EDPB)},
  title = {Statement 3/2024 on data protection authorities' role in the Artificial Intelligence Act framework},
  url = {https://www.edpb.europa.eu/system/files/2024-07/edpb_statement_202403_dpasroleaiact_en.pdf},
  urldate = {2025-07},
  year = {2024}
}

@misc{euartificialintelligenceact_2021,
  title = {The EU Artificial Intelligence Act },
  url = {https://artificialintelligenceact.eu},
  urldate = {2025-08},
  year = {2024},
}

@misc{gdpr_2018_general,
  title = {General Data Protection Regulation (GDPR)},
  url = {https://gdpr-info.eu},
  urldate = {2025-11},
  year = {2018}
}

@inproceedings{donahue2021model,
  title={Model-sharing games: Analyzing federated learning under voluntary participation},
  author={Donahue, Kate and Kleinberg, Jon},
  booktitle={Proceedings of the AAAI Conference on Artificial Intelligence},
  volume={35},
  number={6},
  pages={5303--5311},
  year={2021}
}

@inproceedings{wang2025unlearning,
  title={Unlearning Incentivizes Learning under Privacy Risk},
  author={Wang, Qiyuan and Xu, Ruiling and He, Shibo and Berry, Randall and Zhang, Meng},
  booktitle={Proceedings of the ACM on Web Conference 2025},
  pages={1456--1467},
  year={2025}
}

@article{nguyen2022survey,
  title={A survey of machine unlearning},
  author={Nguyen, Thanh Tam and Huynh, Thanh Trung and Ren, Zhao and Nguyen, Phi Le and Liew, Alan Wee-Chung and Yin, Hongzhi and Nguyen, Quoc Viet Hung},
  journal={arXiv preprint arXiv:2209.02299},
  year={2022}
}

@misc{a2024_ai,
  title = {AI Auditing},
  url = {https://www.edpb.europa.eu/our-work-tools/our-documents/support-pool-experts-projects/ai-auditing_en},
  urldate = {2025-09-08},
  year = {2024},
  organization = {European Data Protection Board (EDPB)}
}

@ARTICLE{ding2025incentivized,
author={Ding, Ningning and Sun, Zhenyu and Wei, Ermin and Berry, Randall},
journal={ IEEE Transactions on Mobile Computing },
title={Incentivized Federated Learning and Unlearning },
year={2025},
volume={24},
number={09},
pages={8794-8810},
}

@misc{lomas_2024_clearview,
  author = {Lomas, Natasha},
  title = {Clearview AI hit with its largest GDPR fine yet as Dutch regulator considers holding execs personally liable},
  url = {https://techcrunch.com/2024/09/03/clearview-ai-hit-with-its-largest-gdpr-fine-yet-as-dutch-regulator-considers-holding-execs-personally-liable/},
  urldate = {2025-09-10},
  year = {2024},
  organization = {TechCrunch}
}

@misc{shrishak_2024_support,
  organization = {European Data Protection Board (EDPB)},
  title = {AI: Complex Algorithms and effective Data Protection Supervision},
  url = {https://www.edpb.europa.eu/our-work-tools/our-documents/support-pool-experts-projects/ai-complex-algorithms-and-effective-data_en},
  urldate = {2025-12-09},
  year = {2025}
}

@misc{a2025_hirundo,
  title = {Hirundo},
  url = {https://www.hirundo.io},
  urldate = {2025-09-24},
  year = {2025},
}

@inproceedings{sarkar2020lethe,
  title={Lethe: A tunable delete-aware LSM engine},
  author={Sarkar, Subhadeep and Papon, Tarikul Islam and Staratzis, Dimitris and Athanassoulis, Manos},
  booktitle={Proceedings of the 2020 ACM SIGMOD International Conference on Management of Data},
  pages={893--908},
  year={2020}
}

@article{xu2024really,
  title={Really unlearned? verifying machine unlearning via influential sample pairs},
  author={Xu, Heng and Zhu, Tianqing and Zhang, Lefeng and Zhou, Wanlei},
  journal={arXiv preprint arXiv:2406.10953},
  year={2024}
}

@article{wang2025evaluation,
  title={Evaluation of Machine Unlearning through Model Difference},
  author={Wang, Weiqi and Zhang, Chenhan and Tian, Zhiyi and Yu, Shui and Su, Zhou},
  journal={IEEE Transactions on Information Forensics and Security},
  year={2025},
  volume={20},
  number={},
  pages={5211-5223},
  publisher={IEEE}
}

@article{zhou2025truvrf,
  title={TruVRF: Towards triple-granularity verification on machine unlearning},
  author={Zhou, Chunyi and Gao, Yansong and Fu, Anmin and Chen, Kai and Zhang, Zhi and Xue, Minhui and Dai, Zhiyang and Ji, Shouling and Zhang, Yuqing},
  journal={IEEE Transactions on Information Forensics and Security},
  year={2025},
  volume={20},
  number={},
  pages={4844-4859},
  publisher={IEEE}
}

@article{xue2025towards,
  title={Towards Reliable Forgetting: A Survey on Machine Unlearning Verification, Challenges, and Future Directions},
  author={Xue, Lulu and Hu, Shengshan and Lu, Wei and Shen, Yan and Li, Dongxu and Guo, Peijin and Zhou, Ziqi and Li, Minghui and Zhang, Yanjun and Zhang, Leo Yu},
  journal={arXiv preprint arXiv:2506.15115},
  year={2025}
}

@misc{amini_2025_swedish,
  author = {Amini, Sina Mindus},
  month = {02},
  title = {Swedish DPA Publishes Annual Report for 2024},
  url = {https://digitalcompliance.snellman.com/swedish-dpa-publishes-annual-report-for-2024/},
  urldate = {2025-11-28},
  year = {2025},
  organization = {EU Digital Compliance Tracker (Snellman)}
}

\end{document}